\documentclass[10pt,twocolumn,letterpaper]{article}
\usepackage{titling}
\usepackage[pagenumbers]{cvpr} % To force page numbers, e.g. for an arXiv version
\usepackage{graphicx}
\usepackage{amsmath}
\usepackage{amssymb}
\usepackage{booktabs}
\usepackage[pagebackref,breaklinks,colorlinks]{hyperref}
\usepackage[capitalize]{cleveref}
\crefname{section}{Sec.}{Secs.}
\Crefname{section}{Section}{Sections}
\Crefname{table}{Table}{Tables}
\crefname{table}{Tab.}{Tabs.}

\usepackage{booktabs}
\usepackage{multirow}
\usepackage{amssymb}
\usepackage{cite}
\usepackage{algorithm}
\usepackage{algorithmic}

\usepackage[table]{xcolor}

\newcommand{\wen}[1]{{\textcolor{black}{#1}}}%blue}{#1}}}
\newcommand{\guo}[1]{{\textcolor{black}{#1}}}

\begin{document}

\title{Multi-Person Extreme Motion Prediction}

\author{
Wen Guo$^{1}$\footnotemark[1], 
Xiaoyu BIE$^{1}$\footnotemark[1], 
Xavier Alameda-Pineda$^1$, 
Francesc Moreno-Noguer$^2$\\
$^1$Inria, Univ. Grenoble Alpes, CNRS, Grenoble INP, LJK, 38000 Grenoble, France\\
$^2$Institut de Robòtica i Informàtica Industrial, CSIC-UPC, Barcelona, Spain\\
$^1${\tt\small \{wen.guo,xiaoyu.bie,xavier.alameda-pineda\}@inria.fr}, $^2${\tt\small fmoreno@iri.upc.edu}}

% \author{
% Wen Guo$^{1}$\footnotemark[1], 
% Xiaoyu BIE$^{1}$\footnotemark[1], 
% Xavier Alameda-Pineda$^1$, 
% Francesc Moreno-Noguer$^2$\\
% $^1$Inria\footnotemark[2]\\
% $^2$Institut de Robòtica i Informàtica Industrial\footnotemark[3]\\
% $^1${\tt\small \{wen.guo,xiaoyu.bie,xavier.alameda-pineda\}@inria.fr}, $^2${\tt\small fmoreno@iri.upc.edu}}

\twocolumn[{%
\renewcommand\twocolumn[1][]{#1}%
\maketitle
\begin{center}
    \centering
    \includegraphics[width=1.0\textwidth]{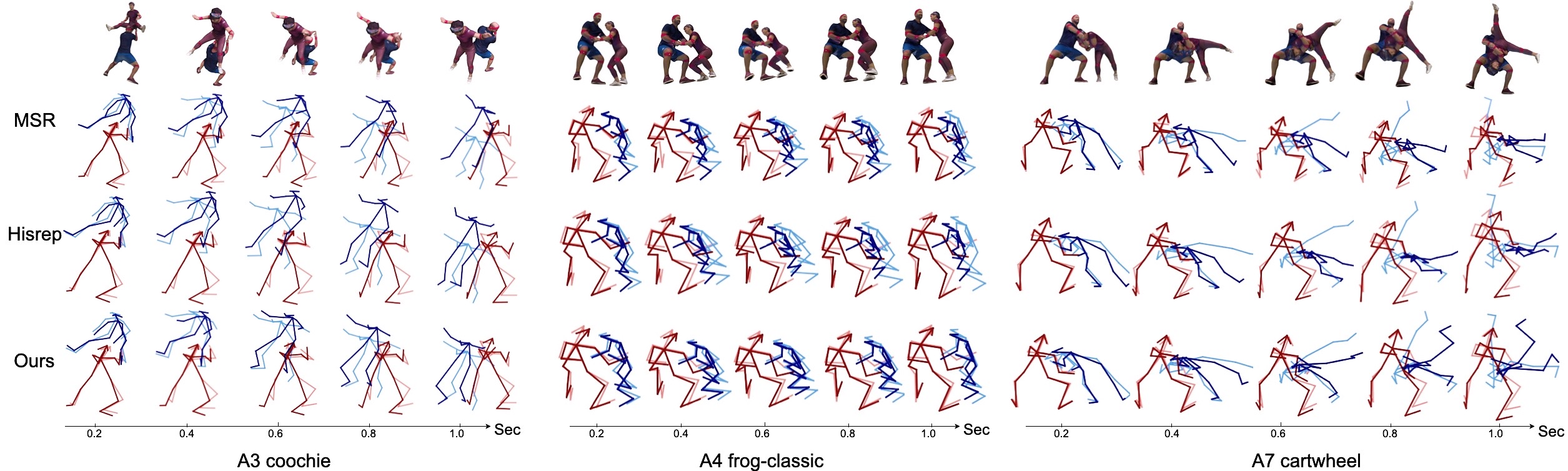}
    \vspace{-5mm}
    \captionof{figure}{{\bf Collaborative human motion prediction. }
    {\bf 1st row:} 3D sample meshes from our ExPI Dataset (just for visualization purposes). {\bf 2nd-4th rows:} Motion prediction results by MSR~\cite{Dang_2021_ICCV}, Hisrep~\cite{mao2020history}, and our method. Dark red/blue indicate prediction results, and light red/blue are the ground truth. By exploiting the interaction information, our approach of collaborative motion prediction achieves significantly better results than methods that independently predict the motion of each person.   %We could see that our result is the closest to the groundtruth, and could succeed even on crazy actions where the single-person methods totally fail. More detailed visualization samples comparing with more SOTA models could be found in supplementary material. 
    }
    \label{fig:teaser}
\end{center}
}]

\renewcommand{\thefootnote}{\fnsymbol{footnote}}
\footnotetext[1]{Equal contribution.}
\footnotetext[2]{This research was supported by ANR-3IA MIAI (ANR-19-P3IA-0003), ANR-JCJC ML3RI (ANR-19-CE33-0008-01), H2020 SPRING (funded by EC under GA \#871245), by the Spanish government with the project MoHuCo PID2020-120049RB-I00 and by an Amazon Research Award. We also thank the Kinovis platform at Inria Grenoble, Laurence Boissieux and Julien Pansiot at Inria Grenoble for their help, and Nvidia for hardware donation under the Academic Hardware Grant Program.
And we thank Yuming Du, Yang Xiao, Meng Guo, Anand Ballou, Louis Airale for their kind suggestions and discussions.}
\renewcommand*{\thefootnote}{\arabic{footnote}}
%%%%%%%%% ABSTRACT
\begin{abstract}
\vspace{-3mm}
   Human motion prediction aims to forecast future poses given a sequence of past 3D skeletons. While this problem has recently received increasing attention, it has mostly been tackled for single humans in isolation. In this paper, we explore this problem when dealing with humans performing collaborative tasks, we seek to predict the future motion of two interacted persons given two sequences of their past skeletons.
   We propose a novel cross interaction attention mechanism that exploits historical information of both persons, and learns to predict cross dependencies between the two pose sequences. Since no dataset to train such interactive situations is available, we collected ExPI (Extreme Pose Interaction) dataset, a new lab-based person interaction dataset of professional dancers performing Lindy-hop dancing actions, which contains 115 sequences with 30K frames annotated with 3D body poses and shapes. We thoroughly evaluate our cross interaction network on ExPI and show that both in short- and long-term predictions, it consistently outperforms state-of-the-art methods for single-person motion prediction. 
   Our code and dataset are available at: \url{https://team.inria.fr/robotlearn/multi-person-extreme-motion-prediction/}
\end{abstract}

%%%%%%%%%%%%%%%%%%%%%%%%%%%%%%%%%%%%%%%%
%% BODY TEXT
\section{Introduction}
 
The goal of human motion prediction is to predict future motions from previous observations. 
With the successful development of deep human pose estimation from single image~\cite{3dmppe,guo2021pi,lcr,lcr++,singleshot,kumarapu2020animepose,von2018recovering,mehta2019xnect,benzine2019deep,dabral2019multi,moreno20173d}, motion prediction begins to draw an increasing attention~\cite{corona2020context, mao2020history, aksan2020spatio, fragkiadaki2015recurrent,gui2018adversarial,hernandez2019human, jain2016structural, martinez2017human, barsoum2018hp, li2018transferable, ghosh2017learning, mao2019learning,sahin2019instance,li2020predicting}.
Most existing works formulate motion prediction as a sequence-to-sequence task, where past observations of 3D skeleton data are used to forecast future skeleton movements.
A common denominator of all these approaches is that they treat each pose sequence as an independent and isolated entity: the motion predicted for one person relies solely on her/his past motion. However, in real world scenarios people interact with each other,  and the motion of one person is typically dependent on or correlated with the motion of other people. Thus, we could potentially improve the performance of motion prediction by exploiting such human interaction.

Based on this intuition, in this paper we present a novel task: \textit{collaborative motion prediction}, which aims to jointly predict the motion of two persons strongly involved in an interaction. To the best of our knowledge, previous publicly available datasets for 3D human motion prediction like 3DPW~\cite{von2018recovering} and CMU-Mocap~\cite{mocap2009} that involve multiple persons only include weak human interactions, e.g., talking, shaking hands etc. Here we move a step further and analyse situations where the motion of one person is highly correlated to the other person, which is often seen in team sports or collaborative assembly tasks in factories.

\begin{figure*}[t!]
    \centering
    \includegraphics[width=\columnwidth]{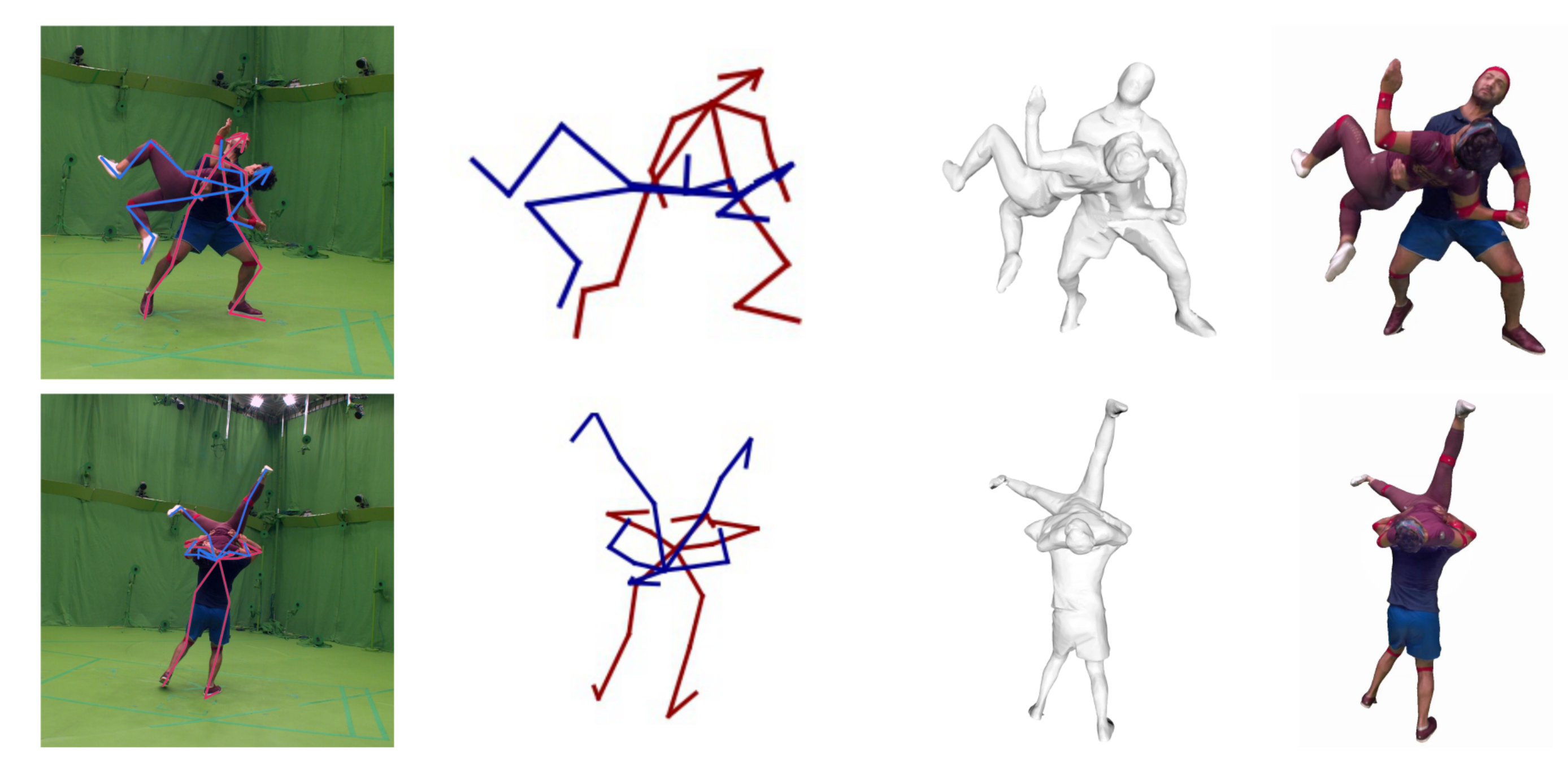}
    \includegraphics[width=\columnwidth]{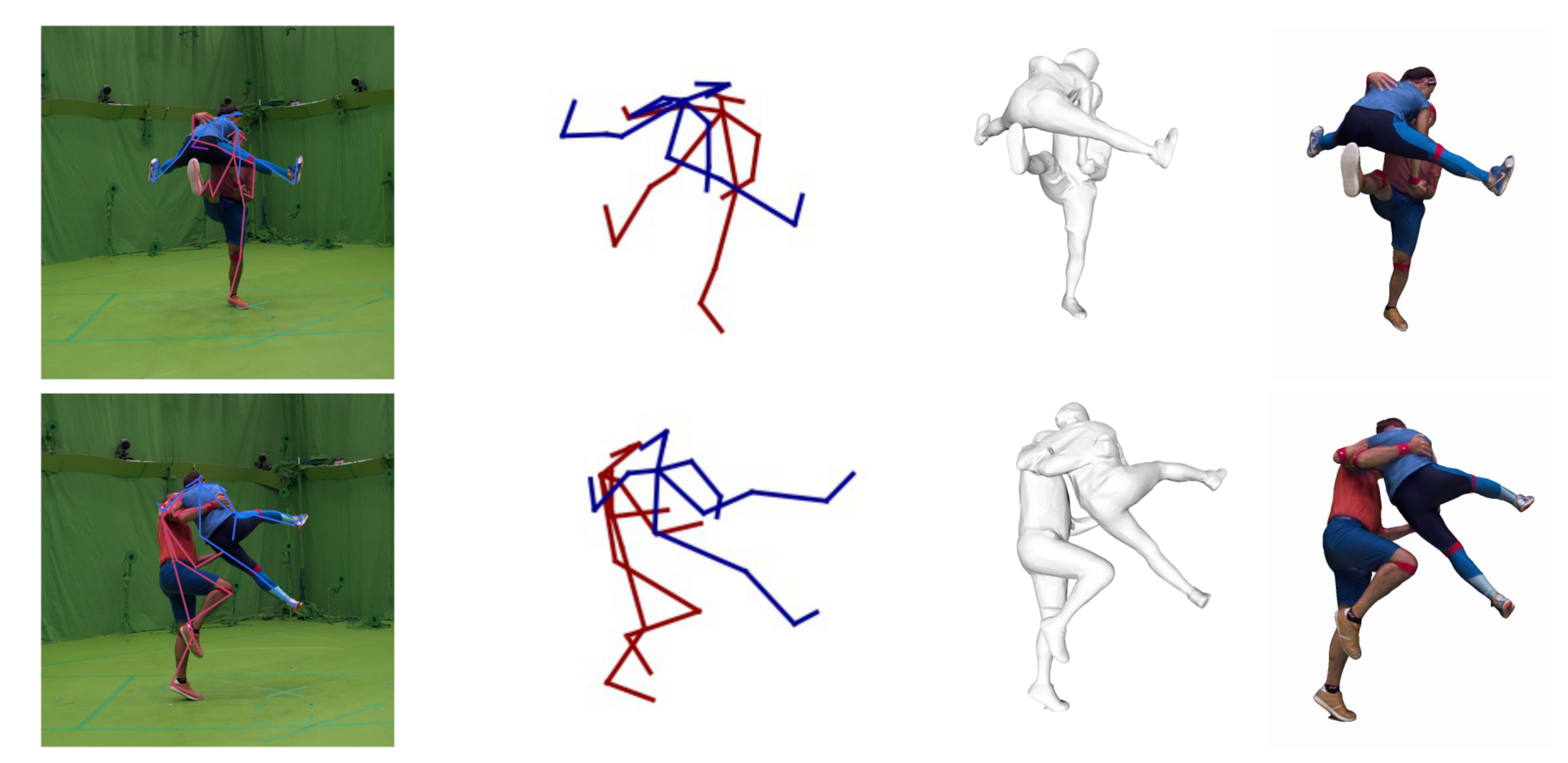}
    \vspace{-3mm}
    \caption{Some samples of the ExPI dataset: RGB image with projected 2D skeletons, 3D pose, mesh and textured mesh.}\vspace{-4mm}
    \label{fig:my_label}
\end{figure*}

With the goal to foster research on this new task, we collected the ExPI (Extreme Pose Interaction) dataset, a large dataset of professional dancers performing Lindy Hop aerial steps.\footnote{The Lindy Hop is an African-American couple dance born in the 1930's in Harlem, New York, see~\cite{monaghan2001study}. } 
To perform these actions, the two dancers perform different movements that require a high level of synchronisation. These actions are composed of extreme poses and require strict and close cooperation between the two persons, which is highly suitable for the study of human interactions.
Some examples of this highly interacted dataset are shown in Figure~\ref{fig:my_label}.
Our dataset contains 115 sequences of 2 professional couples performing 16 different actions. It is recorded in a multiview motion capture studio, and the 3D poses and 3D shapes of the two persons are annotated for all the 30K frames. 
We have carefully created train/test splits, and proposed two different extensions of the pose evaluation metrics for collaborative motion prediction task. 
%We plan to release this dataset to the community. \footnote{The dataset will be released to the community to foster research in this direction after final ethics approval.}

To model such strong human-to-human interactions, we introduce a novel Cross-Interaction Attention (XIA) module,  which is based upon a standard multi-head attention~\cite{vaswani2017attention} and exploits historical motion data of the two persons simultaneously. For a pair of persons engaging in the same activity, XIA module extracts the spatial-temporal motion information from both persons and uses them to guide the prediction of each other.

We exhaustively evaluate our approach and compare it with state-of-the-art methods designed for single human motion prediction. Note that in our dataset of dancing actions, movements are performed at high speed. The long term predictions are very challenging in this case. Nevertheless, the results demonstrate that our approach consistently outperforms these methods by a large margin, with
$10\sim40\%$  accuracy improvement for short ($\leq500$~ms) and $5\sim30\%$  accuracy improvement for long term prediction ($500$~ms $\sim1000$~ms). 

Our key contributions can be summarized as follows:
\vspace{-2mm}
\begin{itemize}
\setlength{\itemsep}{0pt}
\item We introduce the task of collaborative motion prediction, to focus on the estimation of future poses of people in highly interactive setups.
\item We collect and make publicly available ExPI, a large dataset of highly interacted extreme dancing poses, annotated with 3D joint locations and body shapes. We also define the benchmark with carefully selected train/test splits and evaluation protocols.
\item We propose a method with a novel cross-interaction attention (XIA) module that exploits historical motion of two interacted persons to predict their future movements. Our model can be used as a baseline method for collaborative motion prediction.
\end{itemize}
\section{Related Work}

\subsection{3D Human Motion Prediction}
Due to the inherent sequential structure of human motion, 3D human motion prediction has been mostly addressed with recurrent models. For instance, Fragkiadaki~\etal~\cite{fragkiadaki2015recurrent} propose an encoder-decoder framework to embed human poses and an LSTM to update the latent space and predict future motion.
Jain~\etal~\cite{jain2016structural} split human body into sub-parts and forward them via structural RNNs. Martinez~\etal~\cite{martinez2017human} introduce a residual connection to model the velocities instead of the poses themselves. Interestingly, they also show that a model trained with diverse action data performs better than those trained with single actions. However, although RNNs achieve great success in motion prediction, they suffer from containing the entire history with a fixed-size hidden state and tend to converge to a static pose. Some works alleviate this problem by using RNN variants~\cite{liu2019towards, chiu2019action}, sliding windows~\cite{butepage2017deep, butepage2018anticipating}, convolutional models~\cite{holden2015learning, hernandez2019human, li2018convolutional} or adversarial training~\cite{gui2018adversarial}.

Since human body is a non-rigid and structured data, directly encoding the whole body into a compact latent embedding will neglect the spatial connectivity of human joints. To this end, Mao~\etal~\cite{mao2019learning} introduces a feed forward graph convolutional network (GCN)~\cite{kipf2016semi,elickovic2018graph} with learnable adjacent matrix. 
This approach was later boosted with self-attention on an entire piece of historical information~\cite{mao2020history} or a selection of them~\cite{li2021rain}. Recently, GCN based methods are further developed by leveraging multi-scale supervision~\cite{Dang_2021_ICCV}, space-time-separable graph~\cite{Sofianos_2021_ICCV}, and contextual information~\cite{adeli2020socially, Adeli_2021_ICCV}. In terms of GCN design, Cui~\etal~\cite{cui2020learning}  argue that training the adjacent matrix from scratch  ignores the natural connections of human joints, and propose to use a semi-constrained adjacent matrix.  Li~\etal~\cite{Li_2021_ICCV} combine a graph scattering network with a hand-crafted adjacent matrix. Other works also exploit the use of transformers~\cite{vaswani2017attention} to replace GCN in human motion prediction~\cite{cai2020learning,aksan2020spatio}.

Considering that human actions are essentially stochastic in the future, some works leverage on generative models (\eg VAEs and GANs)~\cite{yan2018mt,ZhouLXHH018,aliakbarian2020stochastic,yuan2020dlow,aliakbarian2021contextually,cai2021unified,mao2021generating,petrovich2021action}. Nevertheless, although these models can generate diverse future motions, their prediction accuracy still needs to be further improved when compared to deterministic models.

\subsection{Contextual Information in Human Interaction}

Humans never live in isolation, but perform continuous interactions with other people and objects. Modeling such interactions and the contextual information has been proven to be effective in the topic of 3D human pose estimation~\cite{li2020hmor, hassan2019resolving, zanfir2018sceneconst, wang2019geometric, jiang2020depthordering, guo2021pi}. Contextual information has also been shown to be beneficial in predicting human path trajectories.   For this purpose, recent works explore the use of multi-agent context with social pooling mechanisms~\cite{alahi2016social}, tree-based role alignment~\cite{felsen2018will}, soft attention mechanisms~\cite{vemula2018social} and graph attention networks~\cite{huang2019stgat, kosaraju2019social,li2021spatio}.

Unlike the trajectory forecasting problem that focuses on  a single center point, motion prediction aims at predicting the dynamics of the whole human skeleton. Incorporating contextual information in such a situation is still much unexplored.  
Corona~\etal~\cite{corona2020context} expand the use of contextual information into motion prediction with a semantic-graph model, but only weak human-to-human or human-to-object correlations are modeled. Cao~\etal\cite{cao2020long} involve scene context information into the motion prediction framework, but without human-to-human interaction. More recently, Adeli~\etal\cite{adeli2020socially,Adeli_2021_ICCV} develop a social context aware motion prediction framework, where interactions between humans and objects are modeled either with a social pooling~\cite{adeli2020socially} or with a graph attention network~\cite{Adeli_2021_ICCV}. However, they only study in 2D space~\cite{andriluka2018posetrack} or with weak human interactions~\cite{von2018recovering}. Since in this dataset~\cite{von2018recovering}, most of the actions involve weak interactions like shaking hands or walking together. In any event, none of these papers explores the situation we contemplate in this paper, in which humans do perform highly interactive actions. 

\subsection{Datasets}
Using deep learning methods to study 3D human pose tasks relies on high-quality datasets. Most previous 3D human datasets  are single person~\cite{ionescu2013human3, mahmood2019amass, sigal2010humaneva} or made of pseudo 3D poses~\cite{von2018recovering,singleshotmultiperson2018}. Other datasets which contain lab-based 3D data usually do not have close interactions~\cite{shahroudy2016ntu,liu2019ntu,singleshotmultiperson2018,mocap2009}. Recently, some works start to focus on the importance of context information and propose datasets to model interaction of synthetic persons with scenes~\cite{cao2020long}.
Furthermore, Fieraru~\etal~\cite{fieraru2020CHI3D} created a dataset of human interaction with a contact-detection-aware framework, but this dataset just contains several daily scenarios with mild human interactions and it is not released yet at the time of our submission.
Thus, we believe the ExPI dataset we present here, where the actions of people are highly correlated, fills an empty space in the current datasets of human 3D pose/motion.

\section{Problem Formulation}
\label{sec:task}

As discussed in the introduction, the task of single person human motion prediction is well established. It is defined as learning a mapping $\mathcal{M}: P_{t_{\textsc{i}}:t-1} \xrightarrow{}  P_{t:t_{\textsc{e}}}$ to estimate the future movements $P_{t:t_{\textsc{e}}}$ from the previous observation $P_{t_{\textsc{i}}:t-1}$, where $t_{\textsc{i}}$ ($t_{\textsc{e}}$) denotes the initial (ending) frame of a sequence, and $P_t$ denotes the pose at time $t$.

In this work, we extend the problem formulation to collaborative motion prediction of two interacted persons. While our formulation is general and could work for any kind of interactions, for the sake of consistency throughout the paper, we will denote by $\ell$ and $f$ variables corresponding to the leader and the follower respectively (see Section~\ref{sec:dataset} on the dataset description). Therefore, the collaborative motion prediction task is defined as learning a mapping:
\begin{equation}
\mathcal{M}_{\textsc{c}}: P^\ell_{t_{\textsc{i}}:t-1}, P^f_{t_{\textsc{i}}:t-1} \xrightarrow{}  P^\ell_{t:t_{\textsc{e}}}, P^f_{t:t_{\textsc{e}}}.
\label{equ:collaborative-motion-prediction}
\end{equation}

Since the two persons are involved in the same interaction, we believe it is possible to better predict the motion of a person by exploit the pose information of her/his interacted partner. From now on, we will use $P_t^c = [P_t^l, P_t^f]$ to denote the joint pose of the couple (two actors) at time $t$, and $P_t$ to denote either of them.

\begin{figure*}[t]
\centering
\includegraphics[width=0.99\textwidth]{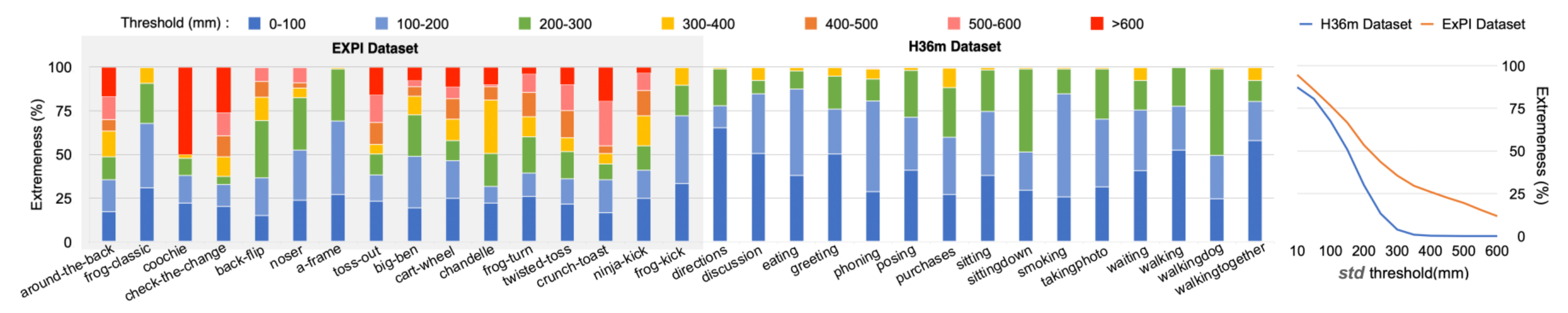}  
  \vspace{-4mm}
  \caption{Extremeness. \textbf{Left}: percentage of joints whose \textit{std} is among a certain threshold (in different colors), for different actions. Actions with more red colors are more extreme. \textbf{Right}: percentage of joints whose \textit{std} is beyond a certain threshold.}\vspace{-5mm}
\label{fig:data_extr_act}
\end{figure*}

In the following parts of the paper, we will provide an experimental framework for the collaborative motion prediction task, consisting of a dataset and evaluation metrics, to foster research in this direction. And we will also introduce our proposed method for this task.

\section{The Extreme Pose Interaction Dataset}
\label{sec:dataset}

We present the Extreme Pose Interaction (ExPI) Dataset, a new person interaction dataset of Lindy Hop dancing actions. In Lindy Hop, the two dancers are called \textit{leader} and \textit{follower}.\footnote{This is the standard gender-neutral terminology for Lindy-Hop.} We recorded 2 couples of dancers in a multi-camera setup equipped also with a motion-capture system. In this section we firstly describe the recording procedure, then give a comprehensive analysis of our dataset.

\subsection{Dataset Overview}
\label{sec:data_structure}

\vspace{1mm}
\noindent\textbf{Dataset Structure.} 
16 different actions are performed in ExPI dataset, some by the 2 couples of dancers, some by only one of the couples. Each action was repeated five times to account for variability. %Overall, ExPI contains 115 sequences, each one depicting an execution of one of the actions. 
More precisely, for each recorded sequence, ExPI provides: (i) Multi-view videos at 25FPS from all the cameras in the recording setup; (ii) Mocap data (3D position of 18 joints for each person) at 25FPS synchronised with the videos.; (iii) camera calibration information; and (iv) 3D shapes as textured meshes for each frame. 
Overall, the dataset contains 115 sequences with 30k visual frames for each view point and 60k 3D instances annotated.

\vspace{1mm}
\noindent\textbf{Dataset Collection and Post-processing.} 
The data were collected in a multi-camera platform
equipped with 68 synchronised and calibrated color cameras and a motion capture system with 20 mocap cameras.\footnote{Kinovis \url{https://kinovis.inria.fr/}} When collecting the motion capture data, some points are missed by the system due to occlusions or tracking losses, which is a common phenomena in lab-based interacted Mocap datasets~\cite{fieraru2020CHI3D}. To overcome this issue and ensure the quality of the data, we spent months to manually label the missing points. \\
More details about the data structure and data post-processing are provided in the supplementary material.

\subsection{Data Analysis}

\vspace{1mm}
\noindent\textbf{Diversity.} 
Similar to Ionescu~\etal~\cite{ionescu2013human3}, we analyse the diversity of our dataset by checking how many \textit{distinct} poses have been obtained. We consider two poses to be \textit{distinct}, if at least one of the $J$ joints for one pose $P^c_{m}$ is different from the corresponding joint of the other pose $P^c_{n}$, beyond a certain tolerance $\tau$ (mm): 
\begin{equation}
    \max_{j\in[1,J]} \lVert P^c_{m,j} - P^c_{n,j} \rVert > \tau,
\end{equation}
where $m, n \in \mathcal{D}$ denote any two poses in the  dataset $\mathcal{D}$.  Then we define \textit{diversity} of the dataset as the percentage of \textit{distinct} poses among all the poses. 
According to Ionescu~\etal~\cite{ionescu2013human3}, the diversity of H3.6M\footnote{Licence for H3.6M dataset \url{http://vision.imar.ro/human3.6m/eula.php}} is $24\%$ and $12\%$ when setting the tolerance $\tau$ to $50$~mm and $100$~mm,  respectively. While the diversities of ExPI for the same threshold values are $52\%$ and $23\%$, which are much more diverse.

\vspace{1mm}
\noindent\textbf{Extremeness.} To measure the extremeness of a pose sequence, we first compute the standard deviation (\textit{std}) over time for each dimension of the $xyz-$coordinate for every joint. %of the three coordinates of every joint . 
Then, the extremeness of the joint $j$ is defined as its maximum per-coordinate standard deviation: $\varepsilon_j = \max\{\sigma_{j}^x,\sigma_{j}^y,\sigma_{j}^z\}.$
Finally, the extremeness of an action is evaluated by computing the percentage of joint extremeness values $\varepsilon_n$ within various intervals $[\varepsilon_{\min},\varepsilon_{\max}]$.
Figure~\ref{fig:data_extr_act} reports the extremeness of ExPI dataset compared to H3.6M in two different ways: (i) a per-action plot reporting extremeness on various color-coded intervals (left); (ii) computing the percentage of joints more extreme than a certain \textit{std} value(right). From both plots it is clear than the ExPI dataset is significantly more extreme than the H3.6M dataset.

\begin{figure*}[t!]
\centering
\includegraphics[width=0.99\linewidth]{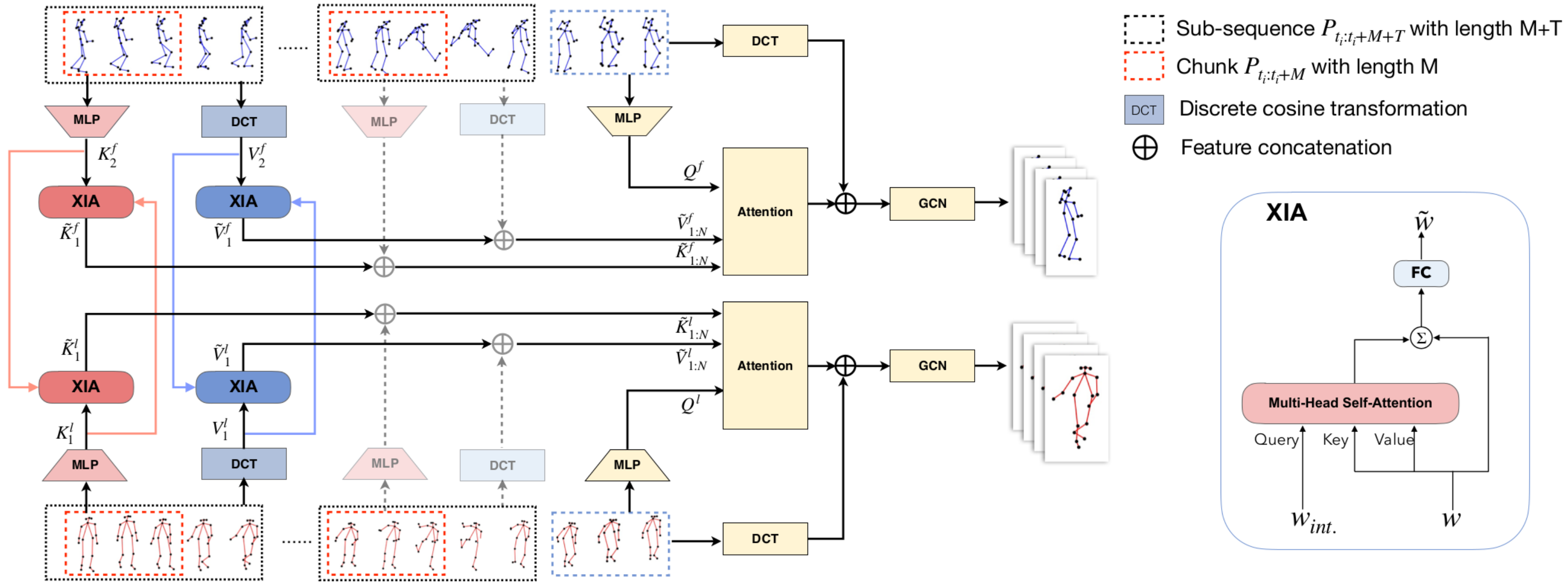}
\vspace{-3mm}
  \caption{\textbf{Left:} Computing flow of the proposed method. Two parallel pipelines -- for the leader and the follower -- are implemented. The key-value pairs are refined by XIA modules (\guo{we just visualize XIA modules for the first sub-sequences, while it is the same for the following sub-sequences}).  \textbf{Right:} Cross-interaction attention (XIA) module. \guo{In order to refine $w$ with the help of the corresponding interaction information $w_{int.}$, the multi-head attention is queried by $w_{int.}$ and take $w$ as key and value.}}\vspace{-5mm}
\label{fig:pipline}
\end{figure*} 

\section{Method}
We introduce our approach for collaborative motion prediction, aiming to set the first performance baseline to help future developments.

\subsection{Pipeline}
\label{sec:pipeline}
The idea of our method is to learn two person-specific motion prediction mappings, and to propose a strategy to share information between these two mappings. The possibility to include information from the other person involved in the interaction, should push the network to learn a better representation for motion prediction.
The overall pipeline is described in Figure~\ref{fig:pipline}-left.

For the two single person motion prediction mappings, we draw inspiration from~\cite{mao2020history}, using an attention model for leaning temporal attention w.r.t. the past motions, and a predictor based on Graph Convolutional Network (GCN)~\cite{kipf2016semi} to model the spatial attention among joints using an adjacency matrix. The temporal attention model aims to find the most relative sub-sequence in the past by measuring the similarity between the last observed sub-sequence and a set of past sub-sequences.
In this attention model, the query $Q$ is learnt by MLP from the last observation $P_{t-1-M:t-1}$  (blue dashed rectangle in Figure~\ref{fig:pipline}-left, length $M$). The keys $K_i$ are learnt by MLP from the starting chunk of sub-sequences $P_{t_{i}:t_{i}+M}$(red dashed rectangles in Figure~\ref{fig:pipline}-left, length $M$).  And the values $V_i$ consist of DCT representations built from the sub-sequences $P_{t_{i}:t_{i}+M+T}$ (black dashed rectangles in Figure~\ref{fig:pipline}-left, length $M+T$), where $t_i$ with $i\in\{1,\ldots,N\}$ indicates the start frame of each past sub-sequence. 

Training such strategy separately for each actor does not account for any interaction between the two dancing partners. To deal with this, we design a cross-interaction attention (XIA) module based on multi-head attention, to introduce guidance form the interacted person. In the next section we introduce this XIA module.

%%%%%%%%%%%%%%%%%%%%%%%%%%%%%%%%%%%%%%%%%%%%%%%%%%%%%%%%%%%%%%%%%%%%%%%% CA split —— AVG L, M
\begin{table*}[t!]
\caption{Results on common action split with the two evaluation metrics (in $mm$). Lower value means better performance. Obviously, our proposal outperforms all the other methods both on JME and AME.}\vspace{-2mm}
\label{tab:proto1}
\centering
\rowcolors{3}{gray!15}{white}
\resizebox{1\textwidth}{!}{\setlength{\tabcolsep}{0.8mm}{
\begin{tabular}{ll|cccc|cccc|cccc|cccc|cccc|cccc|cccc|cccc}
\toprule
\multicolumn{2}{c|}{Action} & \multicolumn{4}{c|}{A1 A-frame} & \multicolumn{4}{c|}{A2 Around the back} & \multicolumn{4}{c|}{A3 Coochie} & \multicolumn{4}{c|}{A4 Frog classic} & \multicolumn{4}{c|}{A5 Noser} & \multicolumn{4}{c|}{A6 Toss Out} & \multicolumn{4}{c|}{A7 Cartwheel} & \multicolumn{4}{c}{AVG} \\
\toprule
\multicolumn{2}{c|}{Time (sec)} & 0.2 & 0.4 & 0.6 & 1.0 & 0.2 & 0.4 & 0.6 & 1.0 & 0.2 & 0.4 & 0.6 & 1.0 & 0.2 & 0.4 & 0.6 & 1.0 & 0.2 & 0.4 & 0.6 & 1.0 & 0.2 & 0.4 & 0.6 & 1.0 & 0.2 & 0.4 & 0.6 & 1.0 & 0.2 & 0.4 & 0.6 & 1.0 \\
\midrule
% \multirow{5}{*}{\rotatebox[origin=c]{90}{\cellcolor{white}\textbf{JME}}}
{\cellcolor{white}} & {\cellcolor{white}Res-RNN~\cite{martinez2017human}}& 83 & 141 & 182 & 236 & 127 & 224 & 305 & 433 & 99 & 177 & 239 & 350 & 74 & 135 & 182 & 250 & 87 & 152 & 201 & 271 & 93 & 166 & 225 & 321 & 104 & 189 & 269 & 414 & 95 & 169 & 229 & 325\\
% &Res. sup.~\cite{martinez2017human}& \\
{\cellcolor{white}} & {\cellcolor{white}LTD~\cite{mao2019learning}} & 70 & 125 & 157 & 189 & 131 & 242 & 321 & 426 & 102 & 194 & 260 & 357 & 62 & 117 & 155 & 197 & 72 & 131 & 173 & 231 & 81 & 151 & 200 & 280 & 112 & 223 & 315 & 442 & 90 & 169 & 226 & 303 \\
{\cellcolor{white}} & {\cellcolor{white}Hisrep~\cite{mao2020history}} & 52 & 103 & {139} & {188} & 96 & 186 & 256 & 349 & 57 & 118 & 167 & 240 & 45 & 93 & 131 & 180 & 51 & 105 & 149 & 214 & 61 & 125 & 176 & 252 & 71 & 150 & 222 & 333 & 62 & 126 & 177 & 251\\
{\cellcolor{white}} &{\cellcolor{white} MSR~\cite{Dang_2021_ICCV}} & 56 & 100 & \textbf{132} & \textbf{175} & 102 & 187 & 256 & 365 & 65 & 120 & 166 & 244 & 50 & 95 & 127 & 172 & 54 & 100 & 138 & 202 & 70 & 132 & 182 & 258 & 82 & 154 & 218 & 321 & 69 & 127 & 174 & 248 \\
\multirow{-5}{*}{\rotatebox[origin=c]{90}{\cellcolor{white}\textbf{JME}}}
& {\cellcolor{white}Ours}  & \textbf{49} & \textbf{98} & 140 & 192 & \textbf{84} & \textbf{166} & \textbf{234} & \textbf{346} & \textbf{51} & \textbf{105} & \textbf{154} & \textbf{234} & \textbf{41} & \textbf{84} & \textbf{120} & \textbf{161} & \textbf{43} & \textbf{90} & \textbf{132} & \textbf{197} & \textbf{55} & \textbf{113} & \textbf{163} & \textbf{242} & \textbf{62} & \textbf{130} & \textbf{192} & \textbf{291} & \textbf{55} & \textbf{112} & \textbf{162} & \textbf{238} \\
%  &new25epo & 50 & 98 & 137 & 183 & 86 & 164 & 226 & 320 & 52 & 106 & 153 & 232 & 43 & 89 & 128 & 173 & 46 & 93 & 135 & 200 & 56 & 114 & 163 & 244 & 65 & 133 & 192 & 280 & 57 & 114 & 162 & 233\\
%  &new30epo & 48 & 95 & 132 & 176 & 85 & 164 & 226 & 325 & 53 & 107 & 156 & 239 & 43 & 88 & 127 & 176 & 45 & 92 & 133 & 197 & 55 & 113 & 161 & 244 & 65 & 134 & 195 & 285 & 56 & 113 & 162 & 235\\

\midrule
% \multirow{5}{*}{\rotatebox[origin=c]{90}{\cellcolor{white}\textbf{AME}}}
{\cellcolor{white}} &{\cellcolor{white}Res-RNN~\cite{martinez2017human}} & 59 & 102 & 132 & 167 & 62 & 112 & 152 & 229 & 57 & 102 & 139 & 215 & 48 & 85 & 113 & 157 & 51 & 90 & 120 & 167 & 53 & 94 & 126 & 183 & 74 & 131 & 178 & 265 & 58 & 102 & 137 & 197\\
% &Res. sup.~\cite{martinez2017human}& \\
{\cellcolor{white}} & {\cellcolor{white}LTD ~\cite{mao2019learning}} & 51 & 92 & 116 & 132 & 51 & 91 & 116 & \textbf{148} & 43 & 80 & 103 & 130 & 38 & 70 & 89 & 111 & 39 & 70 & 90 & 116 & 42 & 75 & 94 & 123 & 52 & 101 & 139 & 198 & 45 & 83 & 107 & 137 \\
{\cellcolor{white}} & {\cellcolor{white}Hisrep~\cite{mao2020history}} & 34 & 69 & 97 & 130 & 44 & 84 & \textbf{115} & 150 & 32 & 65 & 91 & 121 & 27 & 56 & 82 & 112 & 28 & 58 & 85 & 121 & 34 & 66 & 88 & 115 & 42 & 83 & 120 & 171 & 34 & 69 & 97 & 131 \\
{\cellcolor{white}} & {\cellcolor{white}MSR~\cite{Dang_2021_ICCV}}  & 41 & 75 & 99 & \textbf{126} & 54 & 96 & 129 & 180 & 41 & 74 & 98 & 135 & 34 & 61 & 82 & 106 & 33 & 59 & 79 & \textbf{109} & 42 & 71 & 93 & 124 & 57 & 103 & 146 & 210 & 43 & 77 & 104 & 141 \\
\multirow{-5}{*}{\rotatebox[origin=c]{90}{\cellcolor{white}\textbf{AME}}} & {\cellcolor{white}Ours}   & \textbf{32} & \textbf{68} & \textbf{99} & {128} & \textbf{41} & \textbf{82} & 116 & 163 & \textbf{29} & \textbf{58} & \textbf{84} & \textbf{116} & \textbf{24} & \textbf{50} & \textbf{73} & \textbf{96} & \textbf{24} & \textbf{51} & \textbf{75} & \textbf{109} & \textbf{31} & \textbf{62} & \textbf{86} & \textbf{114} & \textbf{41} & \textbf{81} & \textbf{115} & \textbf{160} & \textbf{32} & \textbf{65} & \textbf{93} & \textbf{127}\\
% & new25epo & 34 & 70 & 100 & 126 & 41 & 80 & 111 & 153 & 29 & 60 & 86 & 118 & 25 & 54 & 79 & 103 & 26 & 53 & 77 & 108 & 32 & 62 & 86 & 115 & 40 & 78 & 111 & 154 & 32 & 65 & 93 & 125\\
% & new30epo & 33 & 68 & 97 & 123 & 41 & 80 & 112 & 154 & 29 & 59 & 85 & 115 & 25 & 52 & 77 & 104 & 26 & 53 & 76 & 107 & 31 & 62 & 85 & 115 & 40 & 79 & 111 & 156 & 32 & 65 & 92 & 125\\
 
\bottomrule

\end{tabular}}}
\end{table*}
%%%%%%%%%%%%%%%%%%%%%%%%%%%%%%%%%%%%%%%%%%%%%%%%%%%%%%%%%%%%%%%%%%%%%%%%

\subsection{Cross-Interaction Attention (XIA)}
XIA aims to share motion information between the two predictors. 
In particular, we denote the query and the key-value pairs for one person by $Q$ and  $\{K_i,V_i\}_{i=1}^N$ respectively, and use the superscript $f$ and $\ell$ to indicate the two person, follower and leader. We naturally cast the collaborative human motion prediction task into learning how to jointly exploit the information in $(K_i, V_i)$ when querying with $Q$ to predict motion of each person.

Our intuition is that the pose information (key-value pairs) of one person can be used to transform the pose information of the other person for better motion prediction. We implement this intuition with the help of the proposed \textit{cross-interaction attention module}. Such a module takes as input $w$ and the corresponding vector from the interacted pose $w_{int.}$, and uses multi-head self attention to get the refined vector $\tilde{w}$ (see Figure~\ref{fig:pipline}-right):
\begin{align}
    %\tilde{v} = \textrm{XIA}(v,w) = \textrm{FC}(\textrm{MHA}(w,v,v) + v),
    \tilde{w} & = \textrm{XIA}(w_{int.},w)
     = \textrm{FC}(\textrm{MHA}(w_{int.},w,w) + w),
\end{align} 
where $\textrm{MHA}(q,k,v)$ stands for multi-head attention with query $q$, key $k$ and value $v$, and $\textrm{FC}$ indicates fully connected layers.
We use different XIA modules to update keys and values mentioned in Section~\ref{sec:pipeline}: in our implementation, XIA modules for keys have 8 attention heads, and XIA for values have a single attention head. Moreover, we add a skip-connection for the MHA module followed by 2 FC layers. XIA modules for leader/follower do not share weights.

The proposed XIA module is integrated at several stages of the computing flow as shown in Figure~\ref{fig:pipline}. More precisely, we refine all keys:
\begin{equation}
  \tilde{K}_i^\ell = \textrm{XIA}(K_i^\ell,K_i^f), \qquad \tilde{K}_i^f = \textrm{XIA}(K_i^f,K_i^\ell),
\end{equation}
and analogously for the values. 

XIA could be \guo{potentially} generalised to any number of participants by considering either several XIA modules and fusing their outcome, or performing the fusion at the input of XIA module.

%%%%%%%%%%%%%%%%%%%%%%%%%%%%%%%%%%%%%%%%%%%%%%%%%%%%%%%%%%%%%%%%%%%%%%%%
\begin{figure*}[t]
\centering
\includegraphics[width=0.98\textwidth]{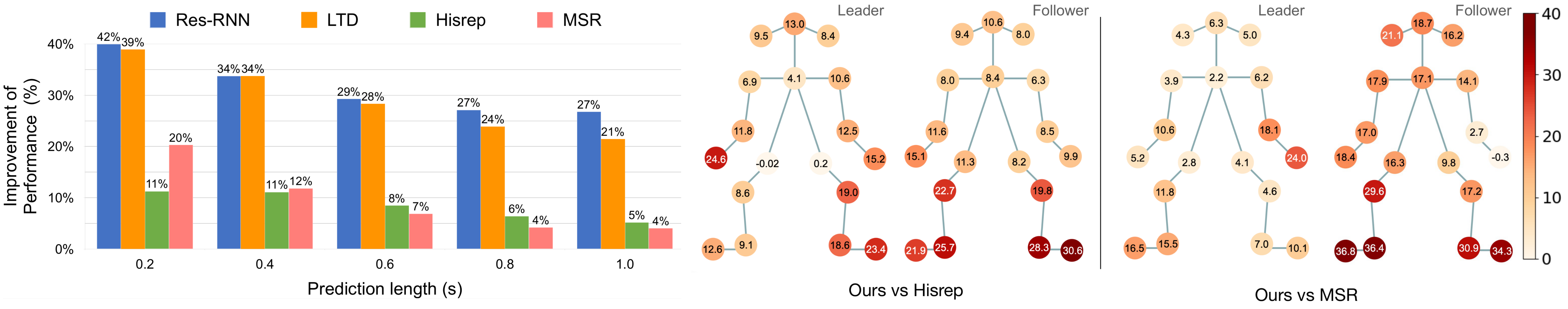}
\vspace{-0.3cm}
  \caption{\textbf{Left:} Percentages of improvement of  our method comparing with different state-of-the-art methods, measured by average JME error on the common action split, at different forecast time. Lower value means closer performance with our model. Our method surpasses these methods up to $10\sim40\%$ on short term,  and $5\sim30\%$ on long term. \textbf{Right:} Joint-wise JME improvement(mm) of our method over Hisrep~\cite{mao2020history} and MSR~\cite{Li_2021_ICCV}. Darker color means larger improvement.}%\wenrmk{(other-our)/other x 100}}
  \vspace{-5mm}
\label{fig:res_ca}
\end{figure*}
%%%%%%%%%%%%%%%%%%%%%%%%%%%%%%%%%%%%%%%%%%%%%%%%%%%%%%%%%%%%%%%%%%%%%%%%

\vspace{7mm}
\subsection{Pose Normalization}
%\vspace{-2mm}
\label{sec:norm}
%Raw poses of ExPI are represented in world coordinate, while the motion prediction task aims at only predicting the relative dynamics related to the root joint, thus we need to normalise the data by removing the global displacement. 
%For single person motion prediction, this could be easily done by defining a coordinate based on a selected root joint.
%We generalize this to the case of multi-person, where we normalize all the people in the same scene by the root joint of one selected person 
Raw poses of ExPI are represented in world coordinate. Similar with single person motion prediction, we normalize the data by removing the global displacement of the poses based on a selected root joint.
While our task aims at predicting not only the the distinct poses but also the relative position of the two person, so 
we have to normalize by the same person to keep the information of their related position. We could either normalize by leader/follower and we choose to normalize by the leader for better visualisation.
Specifically, for each frame, we take the the root joint (middle of the two hips) of the leader as coordinate origin, use the root point and left hip of the leader to define $x$-axis, and use the neck of leader to determine XOZ plane. We normalize all the joints of both persons to this coordinate, then the pose errors can be computed directly in this coordinate. More precisely,
we represent the raw poses in world coordinate as $P_w \in \{{P}^{\ell}_w, {P}^{f}_w\}$, and $T_{{P}^{\ell}_{w,t}}$ is the rigid transformation aligning the two actors to the leader's coordinate system. The normalized coordinates are thus  $P^\ell_t = T_{{P}^{\ell}_{w,t}}{P}^{\ell}_{w,t}$,  and $P^f_t = T_{{P}^{\ell}_{w,t}}{P}^{f}_{w,t}$. In the following $P$ shall always represent the normalized pose unless specified otherwise.

\vspace{-2mm}
\section{Experimental Evaluation}
This section describes the experimental protocol on ExPI, and discuss the results of our proposed method.

%%%%%%%%%%%%%%%%%%%%%%%%%%%%%%%%%%%%%%%%%%%%%%%%%%%%%%%%%%%%%%%%%%%%%%%%%%%%%%%%%%%%%%%%%%%%%%%%%%%% sa split
\begin{table*}[t!]
%\caption{Action-wise results on \textbf{SA} split with the three evaluation metrics (in $mm$). Seven action-wise models are trained independently.}\vspace{-2mm}
\caption{Results on single action split with the two evaluation metrics (in $mm$). Lower value means better performance. Seven action-wise models are trained independently. Our method performs the best in 5 actions, and close to the best for the other 2 actions.}\vspace{-2mm}
\label{tab:proto2}
% \begin{center}
\centering
\rowcolors{3}{gray!15}{white}
\resizebox{0.9\textwidth}{!}{\setlength{\tabcolsep}{0.8mm}{
\begin{tabular}{ll|cccc|cccc|cccc|cccc|cccc|cccc|cccc}
\toprule
\multicolumn{2}{c|}{Action} & \multicolumn{4}{c|}{A1 A-frame} & \multicolumn{4}{c|}{A2 Around the back} & \multicolumn{4}{c|}{A3 Coochie} & \multicolumn{4}{c|}{A4 Frog classic} & \multicolumn{4}{c|}{A5 Noser} & \multicolumn{4}{c|}{A6 Toss Out} & \multicolumn{4}{c}{A7 Cartwheel} \\
\midrule
% \multicolumn{2}{c|}{Frame} & 5 & 10 & 15 & 25 & 5 & 10 & 15 & 25& 5 & 10 & 15 & 25 & 5 & 10 & 15 & 25& 5 & 10 & 15 & 25 & 5 & 10 & 15 & 25& 5 & 10 & 15 & 25\\
\multicolumn{2}{c|}{Time (sec)} & 0.2 & 0.4 & 0.6 & 1.0 & 0.2 & 0.4 & 0.6 & 1.0 & 0.2 & 0.4 & 0.6 & 1.0 & 0.2 & 0.4 & 0.6 & 1.0 & 0.2 & 0.4 & 0.6 & 1.0 & 0.2 & 0.4 & 0.6 & 1.0 & 0.2 & 0.4 & 0.6 & 1.0 \\
% \midrule
\midrule
% \multirow{5}{*}{\rotatebox[origin=c]{90}{\textbf{JME}}}
{\cellcolor{white}} & {\cellcolor{white}Res-RNN~\cite{martinez2017human}} & 75 & 131 & 171 & 226 & 122 & 215 & 287 & 403 & 97 & 174 & 235 & 329 & 73 & 131 & 177 & 246 & 76 & 136 & 184 & 255 & 100 & 184 & 252 & 357 & 88 & 162 & 219 & 293\\
% &Res. sup.~\cite{martinez2017human}& \\
&LTD~\cite{mao2019learning}  & 70 & 126 & 155 & {183} & 131 & 243 & 312 & 415 & 102 & 194 & 252 & 338 & 62 & 117 & 153 & 203 & 71 & 131 & 171 & 231 & 81 & 151 & 199 & 299 & 112 & 223 & 306 & 411 \\
{\cellcolor{white}} & {\cellcolor{white}Hisrep~\cite{mao2020history}} & 66 & {118} & {153} & 190 & 128 & 231 & 308 & 417 & 74 & 143 & 205 & 295 & 64 & 120 & 159 & {191} & 63 & 121 & 166 & 227 & 90 & 168 & 232 & 312 & 88 & 166 & 232 & 332  \\
&MSR\cite{Dang_2021_ICCV}   & \textbf{64} & \textbf{108} & \textbf{136} & \textbf{171} & 119 & 210 & 282 & 385 & 79 & 144 & 189 & 265 & \textbf{59} & \textbf{103} & \textbf{134} & \textbf{173} & 65 & 118 & 162 & 225 & 86 & 151 & 201 & 283 & 96 & 178 & 255 & 362\\
\multirow{-5}{*}{\rotatebox[origin=c]{90}{\cellcolor{white}\textbf{JME}}} 
& {\cellcolor{white}Ours} & \textbf{64} & 120 & 160 & 199 & \textbf{109} & \textbf{200} & \textbf{275} & \textbf{381} & \textbf{59} & \textbf{117} & \textbf{174} & \textbf{277} & {60} & {116} & 162 & 209 & \textbf{53} & \textbf{106} & \textbf{152} & \textbf{221} & \textbf{65} & \textbf{122} & \textbf{166} & \textbf{223} & \textbf{74} & \textbf{144} & \textbf{203} & \textbf{301} \\
% &new25epo  & 67 & 125 & 168 & 205 & 117 & 218 & 294 & 380 & 57 & 119 & 178 & 269 & 60 & 116 & 157 & 200 & 56 & 111 & 156 & 220 & 64 & 119 & 162 & 224 & 79 & 150 & 217 & 322\\

\midrule
% \multirow{5}{*}{\rotatebox[origin=c]{90}{\textbf{AME}}}
&Res-RNN~\cite{martinez2017human}  & 56 & 99 & 129 & 163 & 61 & 110 & 150 & 229 & 53 & 96 & 131 & 188 & 46 & 81 & 106 & 142 & 44 & 79 & 106 & 147 & 53 & 100 & 162 & 176 & 70 & 133 & 163 & 198\\
% &Res. sup.~\cite{martinez2017human}& \\
{\cellcolor{white}} & {\cellcolor{white}LTD~\cite{mao2019learning}} & 51 & 93 & {114} & {127} & \textbf{51} & \textbf{91} & \textbf{116} & \textbf{162} & 43 & 80 & 100 & \textbf{126} & 38 & {70} & \textbf{88} & {118} & 39 & 70 & 90 & 125 & 42 & 75 & \textbf{93} & 123 & 52 & 101 & 137 & 188 \\
&Hisrep~\cite{mao2020history}  & 45 & 83 & 106 & 118 & 57 & 102 & 135 & 178 & 39 & 72 & 100 & 132 & 41 & 77 & 103 & 119 & 35 & 70 & 97 & 125 & 46 & 82 & 107 & 137 & 48 & 90 & 121 & 169 \\
{\cellcolor{white}} & {\cellcolor{white}MSR\cite{Dang_2021_ICCV}} & 46 & 79 & \textbf{98} & \textbf{118} & 60 & 107 & 141 & 192 & 48 & 86 & 111 & 150 & 39 & \textbf{68} & \textbf{88} & \textbf{111} & 39 & 69 & 91 & \textbf{121} & 55 & 93 & 117 & 156 & 66 & 118 & 163 & 222\\
\multirow{-5}{*}{\rotatebox[origin=c]{90}{\textbf{AME}}}
& Ours & \textbf{43} & \textbf{84} & 115 & 131 & 53 & 99 & 136 & 185 & \textbf{35} & \textbf{68} & \textbf{98} & 140 & \textbf{37} & 74 & 106 & 128 & \textbf{29} & \textbf{59} & \textbf{86} & {125} & \textbf{39} & \textbf{72} & 94 & \textbf{119} & \textbf{43} & \textbf{82} & \textbf{112} & \textbf{152} \\
%& Ours 30-5 & 38 & 80 & 113 & 138 & 48 & 93 & 129 & 173 & 32 & 69 & 100 & 146 & 30 & 64 & 92 & 118 & 26 & 56 & 82 & 121 & 36 & 70 & 95 & 123 & 40 & 78 & 111 & 152 \\
%& Ours 50-5    & 37 & 77 & 106 & 134 & 48 & 93 & 129 & 172 & 40 & 76 & 105 & 146 & 33 & 70 & 101 & 122 & 26 & 56 & 83 & 122 & 36 & 69 & 94 & 124 & 41 & 81 & 114 & 153\\
% &new25epo  & 43 & 85 & 116 & 125 & 53 & 98 & 134 & 177 & 34 & 65 & 93 & 125 & 36 & 71 & 98 & 119 & 32 & 66 & 96 & 134 & 38 & 69 & 90 & 115 & 45 & 86 & 118 & 160\\

\bottomrule
\end{tabular}}}\vspace{-3mm}
% \end{center}
\end{table*}
%%%%%%%%%%%%%%%%%%%%%%%%%%%%%%%%%%%%%%%%%%%%%%%%%%%%%%%%%%%%%%%%%%%%%%%%%%%%%%%%%%%%%%%%%%%%%%%%%%%%

\subsection{Splitting the ExPI Dataset}
As described in Sect.~\ref{sec:data_structure}, we record 16 actions in ExPI dataset. Seven of them are common actions ($A_1$ to $A_7$) which are performed by both of the 2 couples: we denote them as $\mathcal{A}_{c}^{1}$ performed by couple 1 and $\mathcal{A}_{c}^{2}$ by couple 2. The other actions are couple-specific, which are performed only by one couple: we denote the actions performed by couple 1 ($A_8$ to $A_{13}$) as $\mathcal{A}_{u}^{1}$, and actions by couple 2 ($A_{14}$ to $A_{16}$) as $\mathcal{A}_{u}^{2}$. With these notations, we propose three data splits.

\vspace{0mm}
\noindent\textbf{Common action split.}
%\noindent\textbf{Common Aerial (CA) split.}
Similar to \cite{ionescu2013human3}, we consider the common actions performed by different couples of actors as train and test data. More precisely, $\mathcal{A}_{c}^{2}$ is the train dataset and $\mathcal{A}_{c}^{1}$ is the test dataset. Thus, train and test data contain the same actions but performed by different people. 

\vspace{0mm}
\noindent\textbf{Single action split.}
%\noindent\textbf{Single Aerial (SA) split.} 
Similar to~\cite{fragkiadaki2015recurrent,jain2016structural}, we train 7 action specific models separately for each common action, by taking one action from couple 2 as train set and the related one from couple 1 as test set.

\vspace{0mm}
\noindent\textbf{Unseen action split.}
%\noindent\textbf{Extra Aerial (EA) split.}
The train set is the entire set of common actions $\{\mathcal{A}_{c}^1, \mathcal{A}_{c}^2\}$. We regard the extra couple-specific actions $\{\mathcal{A}_{u}^{1}, \mathcal{A}_{u}^{2}\}$ as unseen actions and use them as our test set. Thus the train and test data contain both couples of actors, but the test actions are not used in training.

To sum up, common action split is designed for a single model on different actions, single action split is designed for action-wise models, and unseen action split focuses on testing unseen actions to measure methods generalization.

%%%%%%%%%%%%%%%%%%%%%%%%%%%%%%%%%%%%%%%%%%%%%%%%%%%%%%%%%%%%%%%%%%%%%%%%%%%%%%%%%%%%%%%%%%%
\subsection{Evaluation Metrics}
The most common metric for evaluating 3D joint position in pose estimation and motion prediction tasks is the mean per joint position error $\textrm{MPJPE}(P,G) = \frac{1}{J} \sum_{j=1}^{J} \left\|P_{j} - G_{j}\right\|_2$, where $J$ is the number of joints, $P_{j}$ and $G_{j}$ are the estimated and ground truth position of joint $j$. Based on MPJPE, we propose two different metrics to evaluate the multi-person motion task. 

\vspace{1mm}
\noindent\textbf{Joint mean error (JME):} 
We Propose \textit{Joint Mean per joint position Error} to measure poses of different persons in a same coordinate, and denote it as JME for simplicity:
\begin{equation}
\textrm{JME}(P,G) = \textrm{MPJPE}(P, G),
\label{equ:jmpjpe}
\end{equation}
where $P$ and $G$ are normalized (see  Section~\ref{sec:norm}) prediction and ground truth. JME provides an overall idea for the performance of collaborative motion prediction by considering the two interacted persons jointly as a whole, measuring both the error of poses and the error of their relative position.

\noindent\textbf{Aligned mean error (AME):}
We propose \textit{Aligned Mean per joint position Error} to measure pure pose error without the position bias. 
We first erase the errors on the relative position between the two persons by normalizing the poses independently to obtain $\hat{P},\hat{G}$. However the precision of $\hat{P}$ is importantly influenced by the joints that are used to determine the coordinate (hips and back). To mitigate this effect, we compute the best rigid alignment $T_A$ between the estimated pose and the ground-truth {using Procrustes analysis}~\cite{gower1975generalized}:
\begin{equation}
\textrm{AME}(P,G) = \textrm{MPJPE}(T_A(\hat{P}, \hat{G}), \hat{G}),
\label{equ:ampjpe}
\end{equation}
where $\hat{P} \in [\hat{P}^\ell, \hat{P}^f]$ are independently normalized predictions $\hat{P}^\ell_t = T_{P^\ell_t}P^\ell_{t}$ and $\hat{P}_{t}^f = T_{P^f_{t}}P^f_{t}$, and  $T_P$ is the normalisation transformation computed from the pose $P$ as defined in Section~\ref{sec:norm}. The same calculation is done for the ground truth $\hat{G}$. This normalization is only used for evaluation purpose.

%%%%%%%%%%%%%%%%%%%%%%%%%%%%%%%%%%%%%%%%%%%%%%%%%%%%%%%%%%%%%%%%%%%%%%%% EA (UA)
\begin{table*}[t!]
\caption{Action-wise results on unseen action split with the two evaluation metrics (in $mm$). Lower value means better performance. Our method still performs the best on most of the unseen actions and on the average result.}\vspace{-2mm}
% $\ast$ we use abbreviation due to space limit, full name can be found in the supplementary.
\label{tab:proto3}
\centering
\rowcolors{3}{gray!15}{white}
\resizebox{1\textwidth}{!}{\setlength{\tabcolsep}{0.8mm}{
\begin{tabular}{ll|ccc|ccc|ccc|ccc|ccc|ccc|ccc|ccc|ccc|ccc}
\toprule
% \multicolumn{2}{c|}{Action} & \multicolumn{3}{c|}{A8 Back flip} & \multicolumn{3}{c|}{A9 Big ben} & \multicolumn{3}{c|}{A10 Chandelle} & \multicolumn{3}{c|}{A11 Ctc$^\ast$} & \multicolumn{3}{c|}{A12 Frog-turn} & \multicolumn{3}{c|}{A13 Tt$^\ast$} & \multicolumn{3}{c|}{A14 Ct$^\ast$} & \multicolumn{3}{c|}{A15 Frog-kick} & \multicolumn{3}{c|}{A16 Ninja-kick} & \multicolumn{3}{c}{AVG} \\
\multicolumn{2}{c|}{Action} & \multicolumn{3}{c|}{A8} & \multicolumn{3}{c|}{A9} & \multicolumn{3}{c|}{A10} & \multicolumn{3}{c|}{A11} & \multicolumn{3}{c|}{A12} & \multicolumn{3}{c|}{A13} & \multicolumn{3}{c|}{A14} & \multicolumn{3}{c|}{A15} & \multicolumn{3}{c|}{A16} & \multicolumn{3}{c}{AVG} \\
\midrule
\multicolumn{2}{c|}{Time (sec)} &0.4&0.6&0.8&0.4&0.6&0.8&0.4&0.6&0.8&0.4&0.6&0.8& 0.4&0.6&0.8&0.4&0.6&0.8&0.4&0.6&0.8&0.4&0.6&0.8&0.4&0.6&0.8&0.4&0.6&0.8\\
\midrule
% \multirow{5}{*}{\rotatebox[origin=c]{90}{\textbf{JME}}}
{\cellcolor{white}} & {\cellcolor{white}Res-RNN~\cite{martinez2017human}} & 239 & 312 & 371 & 193 & 256 & 303 & 189 & 257 & 310 & 305 & 425 & 520 & 215 & 289 & 348 & 165 & 214 & 252 & 214 & 293 & 357 & 149 & 187 & 210 & 167 & 226 & 277 & 204 & 273 & 327\\
&LTD ~\cite{mao2019learning} & 239 & 324 & 394 & 175 & 226 & 259 & 148 & 191 & 220 & 176 & 240 & 286 & 143 & 178 & 192 & 146 & 193 & 226 & 252 & 333 & 387 & 174 & 228 & 264 & 139 & 184 & 217 & 177 & 233 & 272\\
%&LTD 10-25 ~\cite{mao2019learning} & 204 & 281 & 347 & 157 & 203 & 237 & 134 & 176 & 208 & 151 & 208 & 253 & 130 & 165 & 188 & 124 & 164 & 195 & 207 & 274 & 322 & 162 & 211 & 245 & 131 & 178 & 217 & 156 & 207 & 246 \\
{\cellcolor{white}} & {\cellcolor{white}Hisrep~\cite{mao2020history}} & 195 & \textbf{283} & \textbf{358} & 121 & 169 & 206 & 92 & 129 & \textbf{160} & 129 & 193 & 245 & \textbf{80} & \textbf{104} & 121 & 112 & 154 & 187 & 157 & 219 & 257 & 134 & 190 & 233 & \textbf{96} & \textbf{146} & \textbf{187} & 124 & 176 & \textbf{218}\\
% &MSR\cite{Dang_2021_ICCV}  & 297 & 368 & 451 & 250 & 317 & 395 & 173 & 231 & 303 & 241 & 335 & 416 & 280 & 345 & 449 & 158 & 195 & 246 & 173 & 231 & 289 & \textbf{95} & \textbf{117} & \textbf{134} & 153 & 216 & 268 & 202 & 261 & 327\\
&MSR~\cite{Dang_2021_ICCV} & 230 & 289 & 335 & 188 & 245 & 290 & 148 & 198 & 248 & 234 & 319 & 384 & 176 & 232 & 278 & 162 & 218 & 266 & 177 & 239 & 295 & 143 & 179 & 213 & 157 & 222 & 281 & 179 & 238 & 288 \\
\multirow{-5}{*}{\rotatebox[origin=c]{90}{\cellcolor{white}\textbf{JME}}}
& {\cellcolor{white}Ours}  & \textbf{191} & 287 & 377 & \textbf{118} & \textbf{165} & \textbf{203} & \textbf{91} & \textbf{129} & 162 & \textbf{122} & \textbf{183} & \textbf{232} & 81 & 107 & \textbf{128} & \textbf{106} & \textbf{150} & \textbf{185} & \textbf{156} & \textbf{216} & \textbf{256} & \textbf{126} & \textbf{175} & \textbf{213} & \textbf{96} & 152 & 205 & \textbf{121} & \textbf{174} & \textbf{218}\\
%& Ours 30-5  & 183 & 273 & 358 & 115 & 162 & 197 & 90 & 129 & 163 & 117 & 179 & 234 & 82 & 110 & 135 & 105 & 150 & 188 & 157 & 220 & 264 & 128 & 182 & 225 & 95 & 152 & 209 & 119 & 173 & 219 \\
%& Ours 50-5   & 181 & 268 & 342 & 112 & 156 & 189 & 93 & 135 & 173 & 115 & 176 & 230 & 82 & 110 & 133 & 108 & 157 & 195 & 148 & 212 & 261 & 127 & 178 & 219 & 97 & 158 & 219 & 118 & 172 & 218\\
% &new20epo  & 194 & 292 & 382 & 116 & 162 & 200 & 93 & 130 & 162 & 121 & 181 & 231 & 80 & 105 & 125 & 112 & 156 & 191 & 153 & 212 & 252 & 127 & 178 & 216 & 98 & 156 & 213 & 121 & 175 & 219\\
% &new25epo  & 196 & 299 & 397 & 116 & 161 & 197 & 92 & 130 & 161 & 124 & 187 & 241 & 79 & 102 & 120 & 112 & 155 & 189 & 155 & 214 & 249 & 126 & 175 & 213 & 97 & 155 & 209 & 122 & 175 & 220\\
% &new30epo  & 195 & 298 & 393 & 116 & 162 & 200 & 93 & 131 & 164 & 122 & 182 & 232 & 79 & 103 & 123 & 112 & 157 & 193 & 156 & 215 & 252 & 126 & 177 & 216 & 97 & 157 & 214 & 122 & 176 & 221\\

\midrule
% \multirow{5}{*}{\rotatebox[origin=c]{90}{\textbf{AME}}}
&Res-RNN.~\cite{martinez2017human}& 124 & 165 & 195 & 125 & 157 & 181 & 131 & 166 & 189 & 148 & 198 & 240 & 149 & 169 & 192 & 102 & 128 & 147 & 181 & 237 & 279 & 100 & 129 & 144 & 93 & 124 & 147 & 128 & 164 & 190\\
{\cellcolor{white}} & {\cellcolor{white}LTD ~\cite{mao2019learning}} & \textbf{95} & \textbf{123} & \textbf{146} & 85 & 106 & 116 & 74 & 91 & 101 & 86 & 115 & 137 & 98 & 125 & 134 & 85 & 110 & 124 & 106 & 136 & 155 & 91 & 119 & 135 & 72 & 96 & 116 & 88 & 113 & 129 \\
%&LTD 10-25 ~\cite{mao2019learning} & 84 & 110 & 129 & 77 & 97 & 109 & 67 & 86 & 97 & 76 & 102 & 122 & 89 & 114 & 128 & 72 & 95 & 109 & 89 & 115 & 132 & 86 & 110 & 126 & 69 & 93 & 113 & 79 & 102 & 118 \\
&Hisrep~\cite{mao2020history} & 101 & 144 & 176 & 61 & 82 & 94 & \textbf{49} & \textbf{67} &\textbf{ 80 }& 73 & \textbf{105} & \textbf{129} & \textbf{53 }& \textbf{73 }& \textbf{86} & {64} & {89} & \textbf{104 }& 86 & 120 & \textbf{142} & 73 & 104 & 128 & 54 & 82 & \textbf{104 }& 68 & 96 & \textbf{116}\\
% {\cellcolor{white}} & {\cellcolor{white}MSR\cite{Dang_2021_ICCV}} & 377 & 463 & 315 & 360 & 467 & 308 & 260 & 276 & 212 & 158 & 191 & 211 & 524 & 699 & 344 & 212 & 245 & 167 & 262 & 232 & 230 & \textbf{67} & \textbf{86} & \textbf{98} & 116 & 133 & 142 & 258 & 308 & 225\\
{\cellcolor{white}} & {\cellcolor{white}MSR~\cite{Dang_2021_ICCV}} & 103 & 134 & 155 & 101 & 135 & 160 & 74 & 98 & 121 & 103 & 143 & 173 & 87 & 111 & 132 & 84 & 106 & 122 & 88 & 118 & \textbf{142} & 90 & 113 & 136 & 90 & 122 & 148 & 91 & 120 & 143\\
\multirow{-5}{*}{\rotatebox[origin=c]{90}{\cellcolor{white}\textbf{AME}}}
& Ours & \textbf{95} & 137 & 171 & \textbf{58} & \textbf{80} & \textbf{93} & 51 & 70 & 84 & \textbf{70} & \textbf{105} & 134 & \textbf{53} & \textbf{73} & 88 & \textbf{63} & \textbf{88} & \textbf{104} & \textbf{82} & \textbf{116} & \textbf{142} & \textbf{69} & \textbf{97} & \textbf{120} & \textbf{52} & \textbf{79} & \textbf{104} & \textbf{66} & \textbf{94} & \textbf{116}\\
% %& Ours 30-5   & 91 & 136 & 170 & 59 & 82 & 97 & 50 & 70 & 84 & 66 & 99 & 128 & 54 & 75 & 91 & 61 & 87 & 104 & 83 & 118 & 143 & 69 & 100 & 125 & 52 & 81 & 109 & 65 & 94 & 117\\
% %& Ours 50-5   & 91 & 133 & 165 & 58 & 82 & 98 & 50 & 69 & 83 & 67 & 99 & 125 & 56 & 79 & 96 & 63 & 92 & 113 & 83 & 120 & 146 & 69 & 99 & 123 & 52 & 82 & 109 & 66 & 95 & 118\\
%  &new20epo  & 102 & 151 & 190 & 57 & 77 & 89 & 49 & 67 & 82 & 70 & 102 & 128 & 53 & 73 & 87 & 64 & 90 & 107 & 84 & 118 & 142 & 71 & 100 & 124 & 55 & 86 & 117 & 67 & 96 & 118\\
% &new25epo  & 99 & 145 & 179 & 57 & 77 & 88 & 49 & 68 & 82 & 70 & 103 & 130 & 53 & 72 & 85 & 64 & 91 & 108 & 84 & 118 & 142 & 70 & 98 & 122 & 53 & 83 & 112 & 67 & 95 & 117\\
% &new30epo  & 101 & 148 & 184 & 56 & 76 & 88 & 49 & 68 & 82 & 69 & 101 & 127 & 52 & 72 & 86 & 63 & 89 & 106 & 83 & 117 & 142 & 70 & 100 & 125 & 53 & 83 & 112 & 66 & 95 & 117\\

\bottomrule
\end{tabular}}}\vspace{-4mm}
% \end{center}
\end{table*}

%%%%%%%%%%%%%%%%%%%%%%%%%%%%%%%%%%%%%%%%%%%%%%%%%%%%%%%%%%%%%%%%%%%%%%%%%%%%%%%%%%%%%%%%%%%
\subsection{Implementation Details}
Since this is the first time the collaborative motion prediction task is presented in the literature, there are no available methods to compare with. Thus we choose 4 code-released state-of-the-art methods of single person motion prediction~\cite{martinez2017human,mao2019learning,mao2020history,Dang_2021_ICCV}, and implement their released codes\footnote{All the codes we use are under MIT license.} on ExPI dataset. For fair comparison, all these models are trained with 50 frames of input, train/test for the leader and the follower separately.

We train our model for 25 epochs and calculate the average MPJPE loss of 10 predicted frames. As the data is normalized by the leader, the corresponding branch converges faster, so we compensate by exponentially down-weighting the loss of the leader with the number of epochs $\epsilon$, using the loss function: $\mathcal{L} = \mathcal{L}_{f} + 10^{-\epsilon}\mathcal{L}_{l},$.

When predicting longer horizons, we use the predicted motion as input to predict future motion. {Inspired by~\cite{mao2020history}, we take 64 sub-sequences for each sequence to reduce the variance of the test results.} Overall, we have $7k$ and $2.3k$ sub-sequences for training and testing respectively in the {common action split} and {the single action split}, and $12k$ / $2.9k$  training/testing samples in the {unseen action split}.

%%%%%%%%%%%%%%%%%%%%%%%%%%%%%%%%%%%%%%%%%%%%%%%%%%%%%%%%%%%%%%%%%%%%%%%%%%%%%%%%%%%%%%%%%%%
\vspace{-2mm}
\subsection{Results and Discussion}
\label{sec:res_proto1}

\vspace{-1mm}
\noindent\textbf{Common action split.} 
%\subsubsection{CAsplits} 
Table~\ref{tab:proto1} reports the results  on the \wen{common action split}. %the CA-split. 
We observe that our proposed method outperforms other methods systematically almost for all actions, in all metrics and for different testing time. 
In Figure~\ref{fig:res_ca}-left we calculate the percentage of improvement of our method compared with the state-of-the-art methods, and find that we significantly surpass these methods up to $10\sim40\%$ on short term,  and $5\sim30\%$ on long term. 
We further compare our per-joint results with Hisrep~\cite{mao2020history} and MSR~\cite{Dang_2021_ICCV} in Figure~\ref{fig:res_ca}-right, and observe that our proposed method gets better results on almost on all the joints. More importantly, the keypoints of the limbs (joints of arms and legs) are improved largely. This is reasonable as interaction between persons comes mostly through the limbs, while joints on the torso have little influence on it. So our cross-interaction attention is able to improve the accuracy on the limbs more than on the torso. We could also notice the large improvement on the feet of the follower which usually fly in the air, indicating that our method works even better for these extreme high dynamic joints.  %Hence, it is reasonable that using the cross-interaction gives the most significant boost in the limb joints. 

\noindent\textbf{Single action split and unseen action split.} 
%\subsubsection{EA and SA splits} 
We also reported our proposed method by reporting the results on \wen{single action split and unseen action split}. %the SA and EA splits.
For \wen{single action split},  XIA outperforms the state-of-the-art methods also on action-specific models, as shown in Table~\ref{tab:proto2}. Interestingly, we observe that the performance on \wen{single action split} is worse than the corresponding results on \wen{common action split}, meaning that training on different actions helps regularising the network for this very challenging collaborative extreme motion prediction task.
Regarding \wen{unseen action split} shown in Table~\ref{tab:proto3}, we can see that XIA still outperforms the state-of-the-art methods on most of the actions, demonstrating the robustness of our method.

%%%%%%%%%%%%%%%%%%%%%%%%%%%%%%%%%%%%%%%%%%%%%%%%%%%%%%%%%%%%%%%%%%%%%%%% ABLA w/o SME
\begin{table}[t!]
\caption{Ablations. 
\textit{'mix /cat /sep'} use the single person motion prediction model (Hisrep~\cite{mao2020history}) for multi-person by: mixing two poses together / concatenate two poses as a single vector / train two person-specific models.
%\textit{Sing-cat:} train on concatenation of the two poses using Hisrep;
%\textit{Sing-sep:} train/test leader and follower separately using Hisrep;
\textit{'w.o. XIA'} indicates training leader and follower in parallel using our defined loss without XIA module;
\textit{'XIA kqv / kq / kv / v'} use XIA module to update key, value and query of the temporal attention, or just some of them.
%\textit{'XIA-d kv'} after detaching the query which comes from another person, XIA still works well.
}\vspace{-2mm}
\label{tab:abla}
\centering
\rowcolors{3}{gray!15}{white}
\resizebox{0.48\textwidth}{!}{\setlength{\tabcolsep}{2.5mm}{
\begin{tabular}{cc|ccccc|ccccc}%[cell-space-limits=3pt]
\toprule
\multicolumn{2}{c|}{} & \multicolumn{5}{c|}{\bf JME} & \multicolumn{5}{c}{\bf AME} \\
\midrule
\multicolumn{2}{c|}{Time (sec)} & 0.2 & 0.4 & 0.6 & 0.8 & 1.0 & 0.2 & 0.4 & 0.6 & 0.8 & 1.0 \\
\midrule
{\cellcolor{white}} & {\cellcolor{white}mix} & 69 & 132 & 185 & 233 & 271  & 41 & 77 & 104 & 126 & 142\\
& cat & 61 & 123 & 176 & 223 & 262 & 37 & 71 & 99 & 121 & 138\\
{\cellcolor{white}}& {\cellcolor{white}sep} & 62 & 126 & 177 & 218 & 251 & 34 & 69 & 97 & 116 & 131 \\
\midrule
& w.o. XIA  & 58 & 120 & 174 & 217 & 249 & 33 & 68 & 98 & 118 & 131\\
{\cellcolor{white}} & {\cellcolor{white}XIA kq} & 58 & 118 & 169 & 211 & 245 & 33 & 67 & 95 & 114 & 128\\
& XIA kqv   & 57 & 117 & 170 & 215 & 251 & 32 & 65 & 95 & 116 & 131\\
{\cellcolor{white}} & {\cellcolor{white}XIA v} & 56 & 116 & 168 & 210 & 244 & 32 & 66 & 94 & 113 & \textbf{127}\\
& XIA kv  & \textbf{55} & \textbf{112} & \textbf{162} & \textbf{204} & \textbf{238} & \textbf{32} & \textbf{65} & \textbf{93} & \textbf{112} & \textbf{127}\\
%\midrule
%& XIA-d kv & 56 & 115 & 166 & 208 & 242 & 32 & 66 & 94 & 113 & 126\\
\bottomrule
\end{tabular}}}\vspace{-6mm}
\end{table}

\noindent\textbf{Qualitative results.} 
Figure~\ref{fig:teaser} shows some example of our visualisation results compared to  Hisrep~\etal~\cite{mao2020history}, MSR~\cite{Dang_2021_ICCV} and the ground truth, on the \wen{common action split}. We can see that the poses estimated by our method are much closer to the ground truth than the other methods, and it works well even on some extreme actions where other methods totally fail (Figure~\ref{fig:teaser}-right). More qualitative examples could be found in the supplementary material.

\noindent\textbf{Ablation study.} 
Taking Hisrep~\cite{mao2020history} as example, we first tried 3 different ways of training the single-person motion prediction models on our multi-person dataset: (i) 'mix': train a single model using data of the two poses $\{P^l, P^f\}$; (ii) 'cat': concatenate the two poses as a single input vector $[P^l, P^f]$; (iii) 'sep': train two person-specific models for ${P^l}$ and ${P^f}$. Since 'sep' gives best performance, all the state-of-the-art methods reported above in this paper is using this setting.
As for our collaborative motion prediction model, we report performances of several different design choices of our model. We found that updating the key and values of the temporal attention using our XIA module provide the best results.
We demonstrate the interest of the design of our method as the proposed one is the best in performance and our method significantly improves all the single-person motion prediction results.

%% conclusion
\noindent\textbf{Limitations.}
Collecting clean and reusable 3D pose data requires specific equipment and recording extreme poses requires actors with specific skills, thus ExPI is rare and difficult to reproduce/extend. This is clearly a limitation in the era of data-hungry deep learning architectures. Besides, predicting very long future (beyond~2s for example) is still an open problem, specially for the fast movements of ExPI.

\vspace{-2mm}
\section{Conclusion}
\vspace{-2mm}
Current motion prediction methods are restricted to single person. We move beyond existing approaches for 3D human motion prediction by considering a scenario with two persons performing highly interactive activities. We collected a new dataset called ExPI of professional actors performing dancing actions. ExPI is annotated with sequences of 3D body poses and shapes, opening the door to not only being applied for interactive motion prediction but also for single-frame pose estimation or multi-view 3D reconstruction. In order to learn the interacted motion dynamics, we introduce a baseline method trained with ExPI that exploits historical information of both people in an attention-like fashion. Results of our method show consistent improvement compared to methods that independently predict the motion of each person. 

%%%%%%%%% REFERENCES
{\small
\bibliographystyle{ieee_fullname}
\bibliography{egbib}
}

% Appendix
% \newpage

% \title{Multi-Person Extreme Motion Prediction: Supplementary Material}
% \maketitle
% \setcounter{page}{1}
\setcounter{section}{0}
\renewcommand{\thesection}{\Alph{section}}
%%%%%%%%%%%%%%%%%%%%%%%%%%%%%%%%%%%%%%%%%%%%%%%%%%%%%%%%%%%%%%%%%%%%%%%%%%%%%%%%%%%%%%%%%%%%%%%%%%%%%%%%%%%%%%%%%%%%%%%%%%%%%%%%%%%%%%%%%%%%%%%%%%%%%%%%%%%%%%%%

\section*{Supplementary Material}
\section{Personal data/Human subjects}

Our data collection strategy went through an Ethics Review Board, and the recordings where authorised, together with the associated Consent Form. Our data does not contain any personally identifiable information beyond the images themselves. The data will be shared respecting all national and international regulations, as authorised by COERLE, the Ethics Review Board at INRIA.

\section{More information about the dataset}

\subsection{Data Post-processing}
As introduced in the main paper, it is a common phenomena in lab-based interaction Mocap datasets that many points are missing due to occlusions or tracking loss. This is even worse when dealing with extreme poses. To overcome this we have designed and implemented a 3D hand labelling toolbox.

For each missed value, we choose two orthogonal views among the several viewpoints, and label the missed keypoints by hand on these two frames to get two image coordinates. We then use the camera calibration to back project these two image coordinates into the 3D world coordinate, obtaining two straight lines. Ideally, the intersection of these two lines is the world coordinate of this missing point. Since these two lines do not always intersect, we find the nearest point, in the least-squares sense, to these two lines to approximate the intersection. 

\begin{figure}[t]
\centering
\includegraphics[width=0.8\linewidth]{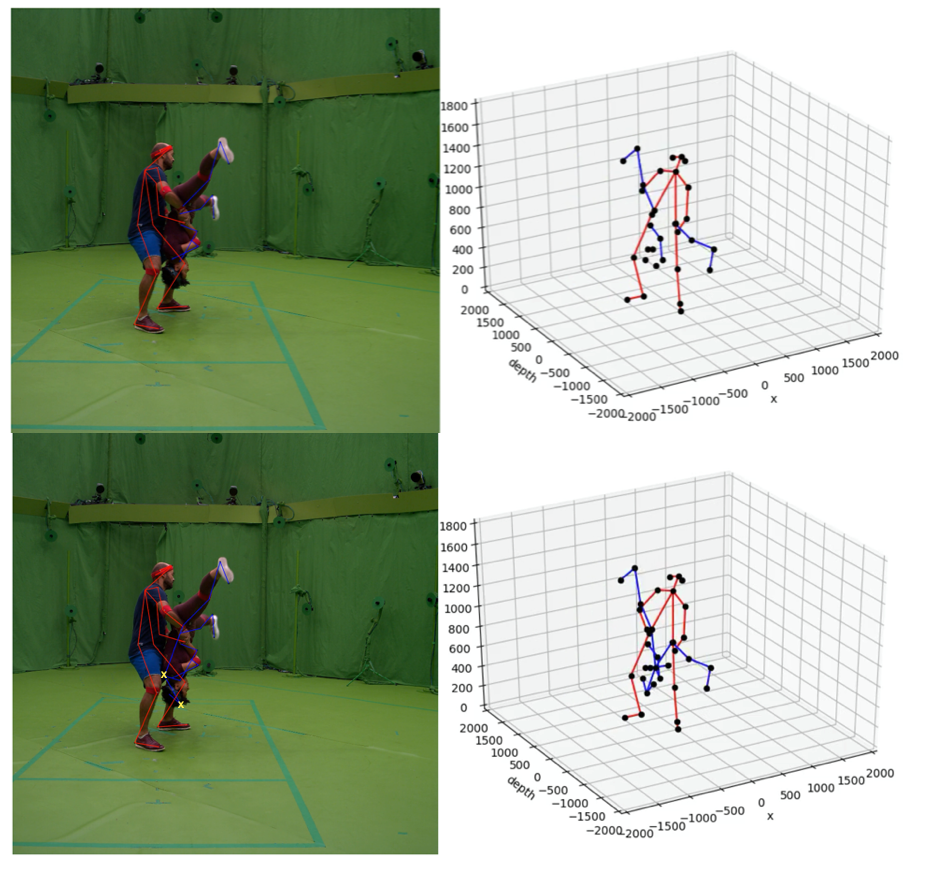}
  \caption{Data-cleaning. \textbf{Top:}Data before cleaning. The two joints 'F-back' and 'F-fhead' are missed. \textbf{Bottom:} Data after cleaning. The yellow marks indicates the two relabeled joints.}\vspace{-3mm}
\label{fig:data_clean}
\end{figure}

In this procedure we did not use the distortion parameters, since we observed that the distortion error is negligible on the views we choose for labeling. The intersection is projected into 3D and various 2D images to confirm the quality of the approximation by visual inspection. 
Figure~\ref{fig:data_clean} shows an example of labeling the missing joints.

\subsection{Action names and joint order}
Table~\ref{tab:action_names} shows the name of the 16 actions performed by the couples of actors in ExPI.
In the video of the supplementary material, we include example videos for each of the 16 actions. 
\begin{table}
\caption{Composition of the ExPI Dataset. The seven first actions are performed by both couples. Six more actions are performed by Couple 1, while three others by Couple 2.}
\centering
\resizebox{.35\textwidth}{!}{
    \setlength{\tabcolsep}{1mm}{
    \begin{tabular}{ l l c c }
    \toprule
    Action & Name & Couple 1 & Couple 2 \\
    \midrule
    $A_1$ & {A-frame} & $\checkmark$ & $\checkmark$ \\
    $A_2$ & {Around the back} & $\checkmark$ & $\checkmark$ \\
    $A_3$ & {Coochie} & $\checkmark$ & $\checkmark$ \\
    $A_4$ & {Frog classic} & $\checkmark$ & $\checkmark$ \\
    $A_5$ & {Noser} & $\checkmark$ & $\checkmark$ \\
    $A_6$ & Toss out & $\checkmark$ & $\checkmark$ \\
    $A_7$ & Cartwheel & $\checkmark$ & $\checkmark$ \\\midrule
    $A_{8}$ &  Back flip & $\checkmark$ & \\
    $A_{9}$ &  Big ben & $\checkmark$ & \\
    $A_{10}$ & Chandelle & $\checkmark$ & \\
    $A_{11}$ & Check the change & $\checkmark$ & \\
    $A_{12}$ & Frog-turn & $\checkmark$ & \\
    $A_{13}$ & Twisted toss & $\checkmark$ & \\\midrule
    $A_{14}$ & Crunch-toast & & $\checkmark$ \\
    $A_{15}$ & Frog-kick & & $\checkmark$ \\
    $A_{16}$ & Ninja-kick & & $\checkmark$ \\
    \bottomrule
    \end{tabular}}
}
\label{tab:action_names}
\end{table}
In the ExPI dataset, the pose of each person is annotated with 18 keypoints, so we have 36 keypoints for both actors. The order of the keypoints is as follows, where ``F'' and ``L'' denote the Follower and the Leader respectively, and ``f'', ``l'' and ``r'' denote ``forward'', ``left'' and ``right'':
\begin{table}[H]
    \centering
    \vspace{-2mm}
    \begin{tabular}{lll}
        (0) `L-fhead' & (1) `L-lhead' & (2) `L-rhead' \\
        (3) `L-back' & (4) `L-lshoulder' & (5) `L-rshoulder'\\
        (6) `L-lelbow' & (7) `L-relbow' & (8) `L-lwrist' \\
        (9) `L-rwrist' & (10) `L-lhip' & (11) `L-rhip' \\
        (12) `L-lknee' & (13) `L-rknee' & (14) `L-lheel' \\
        (15) `L-rheel' & (16) `L-ltoes' & (17) `L-rtoes' \\
        (18) `F-fhead' & (19) `F-lhead' & (20) `F-rhead' \\
        (21) `F-back' & (22) `F-lshoulder' & (23) `F-rshoulder' \\
        (24) `F-lelbow' & (25) `F-relbow' & (26) `F-lwrist' \\
        (27) `F-rwrist' & (28) `F-lhip' & (29) `F-rhip' \\
        (30) `F-lknee' & (31) `F-rknee' & (32) `F-lheel' \\
        (33) `F-rheel' & (34) `F-ltoes' & (35) `F-rtoes'
    \end{tabular}
    \label{tab:my_label}
\end{table}

\subsection{Comparison with other datasets}
%\textcolor{orange}{We need to explain which datasets are shown and why.}
Table~\ref{tab:compara_datasets} compares our dataset with several other public available 3D human datasets that are widely used in recent work~\cite{martinez2017human,mao2019learning,mao2020history,Dang_2021_ICCV}. From this table, we can see that our dataset is eminently suitable for the task of multi-person extreme motion prediction, and it is also able to be used in human pose estimation in rare condition and challenging human shape estimation.

% \textcolor{red}{And similar to Figure~3 in the main paper, we compare the extremeness of ExPI with more datasets in Figure~\ref{}.}
% \begin{table}[t!]
% \caption{Comparison of ExPI with other publicly available \wen{lab-based} datasets on 3D human modeling.}
% \centering
% \resizebox{\columnwidth}{!}{
%     \begin{tabular}{l|ccccc}
%     \bottomrule
%     Dataset& AMASS\cite{mahmood2019amass}  &  H3.6m\cite{ionescu2013human3} & MuPoTS\cite{singleshotmultiperson2018} & ExPI \\
%     \midrule
%     3D joints & $\checkmark$ & $\checkmark$ & $\checkmark$ & $\checkmark$\\
%     Video & $\checkmark$ &  $\checkmark$ & & $\checkmark$\\
%     Shape &  $\checkmark$ &  $\checkmark$ & & $\checkmark$\\
%     Multi-person &  & & $\checkmark$  & $\checkmark$\\
%     Extreme poses & $\checkmark$ & & & $\checkmark$\\
%     %Released & $\checkmark$ & $\checkmark$ & $\checkmark$ &  & $\checkmark$\\
%     Multi-view & & & & $\checkmark$\\
%     \bottomrule
%     \end{tabular}
% }
% \label{tab:compara_datasets}
% \end{table}

\begin{table}[t!]
\caption{Comparison of ExPI with other publicly available datasets commonly used for human motion prediction.}
\centering
\resizebox{\columnwidth}{!}{
    \begin{tabular}{l|cccccc}
    \bottomrule
    Dataset& AMASS\cite{mahmood2019amass}  &  H3.6m\cite{ionescu2013human3} & 3DPW\cite{vonMarcard2018} & MuPoTS\cite{singleshotmultiperson2018} & ExPI \\
    \midrule
    3D joints & $\checkmark$ & $\checkmark$ & $\checkmark$ &  $\checkmark$ & $\checkmark$\\
    Video & $\checkmark$ &  $\checkmark$ & $\checkmark$ & $\checkmark$ & $\checkmark$\\
    Shape &  $\checkmark$ &  $\checkmark$ & $\checkmark$ & & $\checkmark$\\
    Multi-person &  & & $\checkmark$  &  $\checkmark$& $\checkmark$\\
    Extreme poses & $\checkmark$ & & & & $\checkmark$\\
    %Released & $\checkmark$ & $\checkmark$ & $\checkmark$ &  & $\checkmark$\\
    Multi-view & & & & & $\checkmark$\\
    \bottomrule
    \end{tabular}
}
\label{tab:compara_datasets}
\end{table}

%%%%%%%%%%%%%%%%%%%%%%%%%%%%%%%%%%%%%%%%%%%%%%%%%%%%%%%
\section{More Qualitative results}
More qualitative results could be found at the end of this file. We compare our model with models that independently predict the motion of each person, i.e. Res-RNN~\cite{martinez2017human}, LTD~\cite{mao2019learning}, Hisrep~\cite{mao2020history} and MSR~\cite{Dang_2021_ICCV}. Our results are much closer to the ground truth, and it works well even on some extreme actions where other methods totally fail.

\begin{figure*}[t!]
\centering
\includegraphics[width=0.9\linewidth]{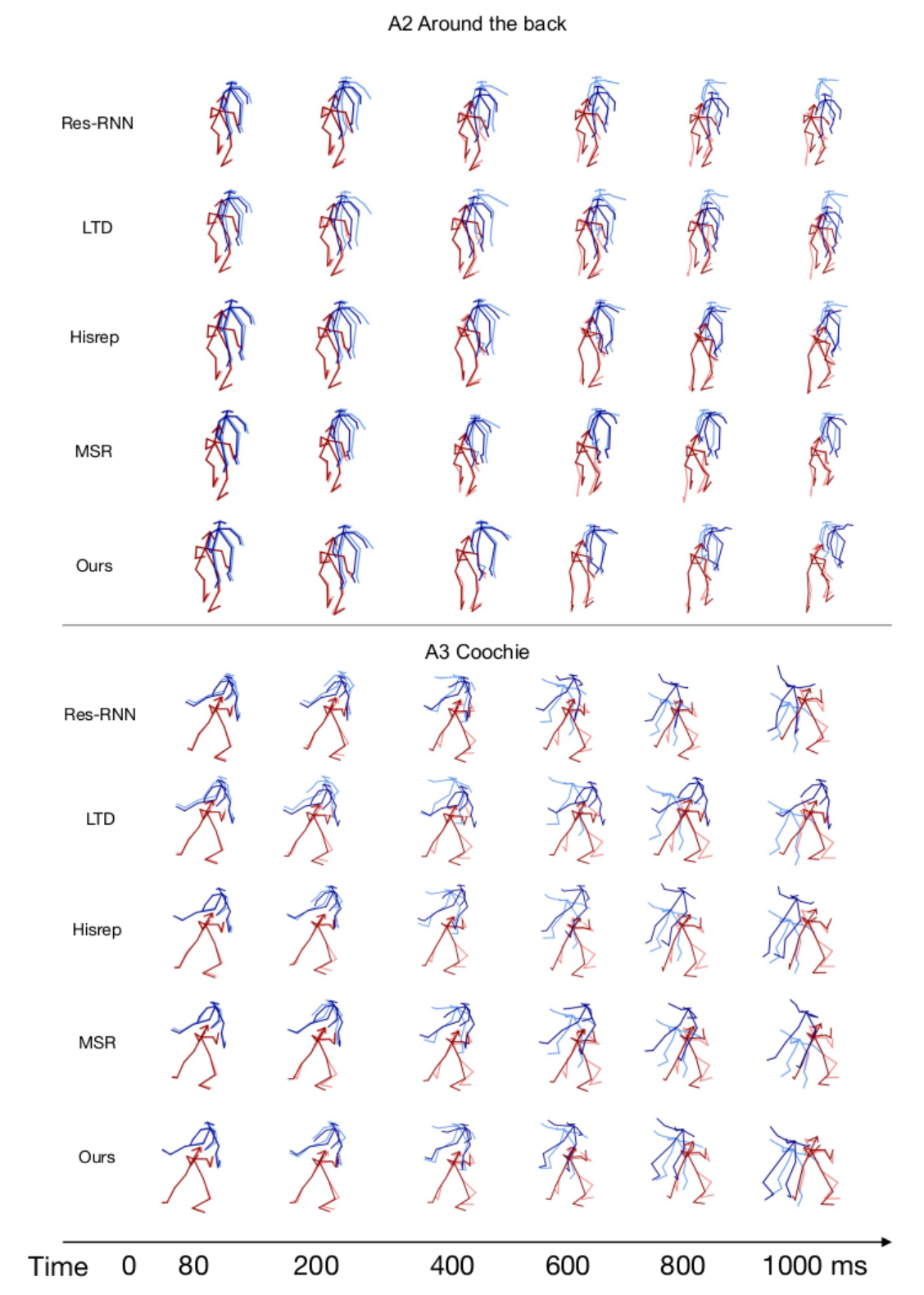}
\vspace{0cm}
%   \caption{some visualization results (draft)\wenrmk{we will add 2 more methods and 3 more examples}}\vspace{-3mm}
\label{fig:res_more_vis_a1}
\end{figure*} 

\newpage

\begin{figure*}[t!]
\centering
\includegraphics[width=0.9\linewidth]{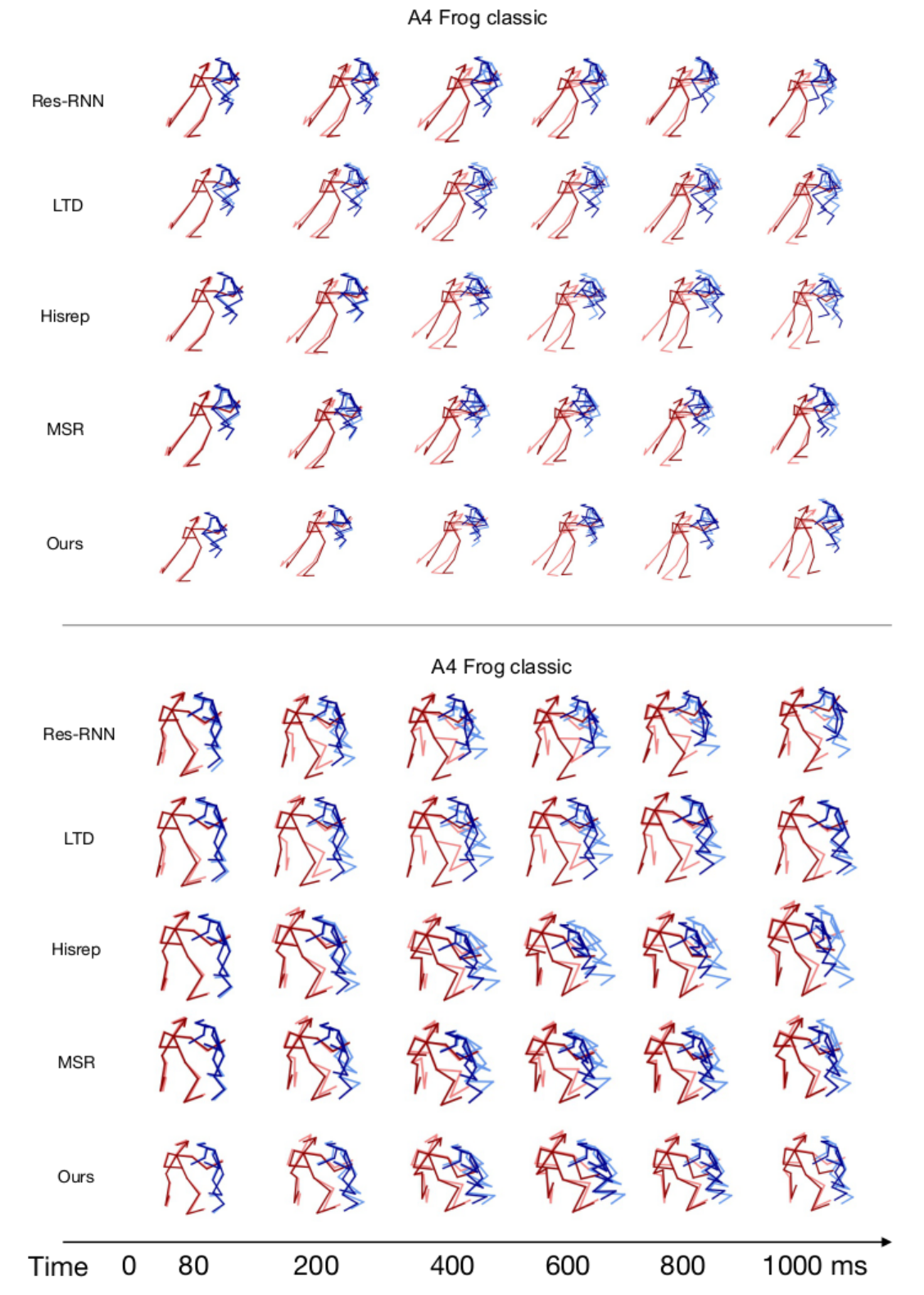}
\vspace{0cm}
%   \caption{some visualization results (draft)\wenrmk{we will add 2 more methods and 3 more examples}}\vspace{-3mm}
\label{fig:res_more_vis_a2}
\end{figure*} 

\newpage

\begin{figure*}[t!]
\centering
\includegraphics[width=0.9\linewidth]{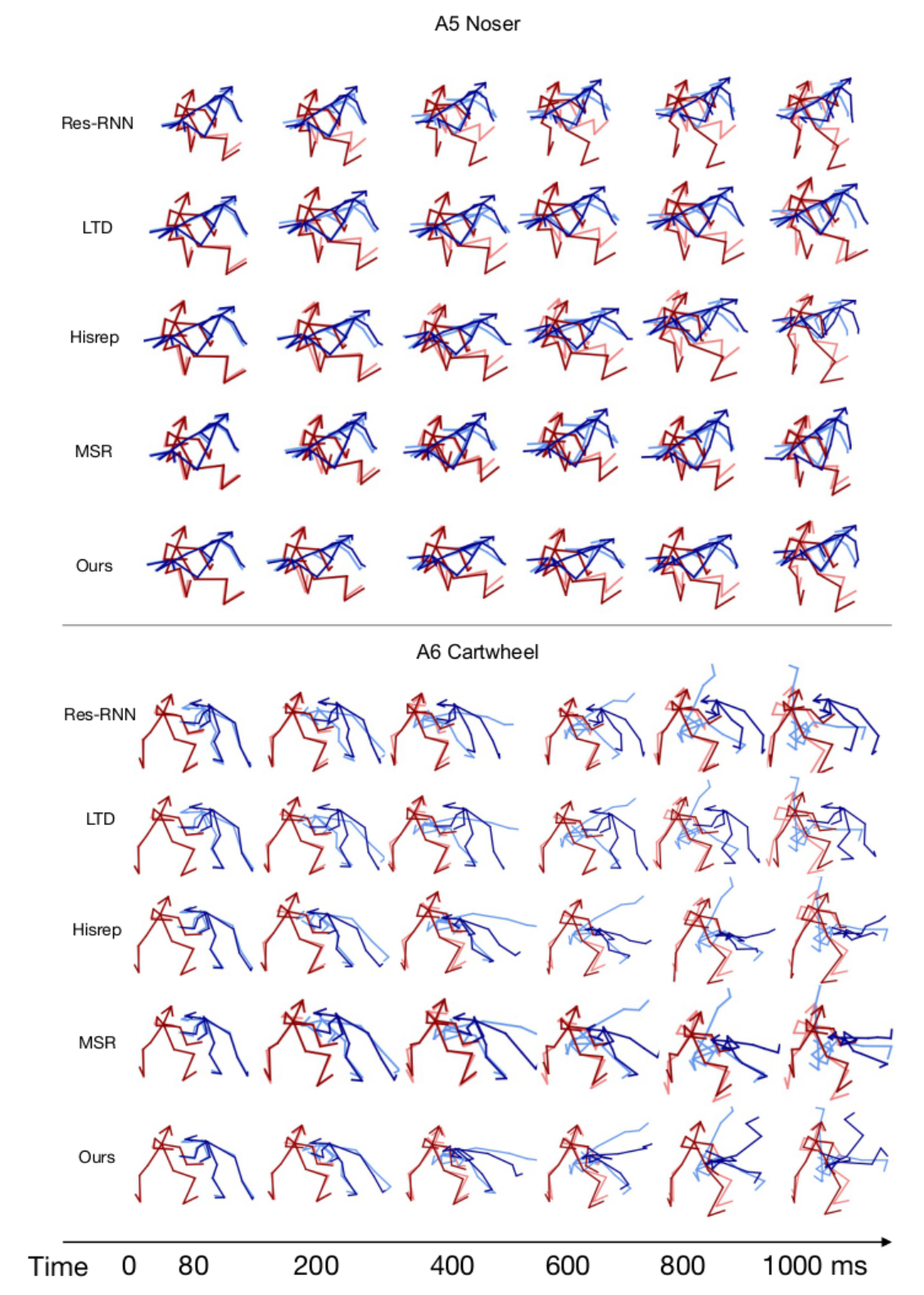}
\vspace{0cm}
%   \caption{some visualization results (draft)\wenrmk{we will add 2 more methods and 3 more examples}}\vspace{-3mm}
\label{fig:res_more_vis_a3}
\end{figure*}

\end{document}